# Evaluating Intelligent Algorithms for Gait Phase Classification in Lower Limb Robotic Systems

Barath Kumar JK and Aswadh Khumar G S

**Abstract—** Accurate and rapid detection of gait phases is of utmost importance in achieving optimal performance of powered lower-limb prostheses and exoskeletons. With the increasing versatility and complexity of these robotic systems, there is a growing need to enhance the performance of gait detection algorithms. The development of reliable and functional gait detection algorithms holds the potential to enhance precision, stability, and safety in prosthetic devices and other rehabilitation technologies. In this systematic review, we delve into the extensive body of research and development in the domain of gait event detection methods, with a specific focus on their application to prosthetic devices. Our review critically assesses various proposed methods, aiming to identify the most effective approaches for gait phase classification in lower limb robotic systems. Through a comprehensive comparative analysis, we highlight the strengths and weaknesses of different algorithms, shedding light on their performance characteristics, applicability, and potential for further improvements. This comprehensive review was conducted by screening two databases, namely IEEE and Scopus. The search was limited to 204 papers published from 2010 to 2023. A total of 6 papers that focused on Heuristic, Thresholding, and Amplitude – Zero Crossing involved techniques were identified and included in the review. 33.3% of implemented Algorithms used kinematic parameters such as joint angles, joint linear and angular velocity, and joint angular acceleration. This study purely focuses on threshold-based algorithms and thus paper focusing on other gait phase detection methods were excluded.

*Index Terms*— event detection; gait phase detection; assistive devices; gait phase classification; lower limb prosthesis; wearable sensors; IMU sensor; EMG

## 1. INTRODUCTION

Human Gait Cycle is a complex process that can be divided into two main phases: stance and swing. Understanding and accurately identifying the different sub-phases within these phases is crucial for evaluating and diagnosing various gait abnormalities. Yi et.al [86] These abnormalities can result from conditions such as brain injuries, which commonly lead to drop foot. Functional Electrical Stimulation (FES) has emerged as a widely accepted method for rehabilitating drop foot. By using FES, the correction of this condition can be achieved effectively. However, to implement FES or any other gait correction technique, it is essential to detect specific gait events accurately. Rueterbories et. Al [14] Traditionally, force-based measurement systems like footswitches or force sensitive resistors (FSRs) have been considered as the standard for gait event detection. Zakria et.al [18] These systems can reliably capture information about initial contact (IC) and foot off contact (FO) events, which are crucial for calculating important gait parameters like stride length, cadence, and single and double support time percentages. However, force-based systems have some limitations. They are prone to mechanical failure, making them less reliable in long-term use. Additionally, when utilized by patients with drop foot, these systems can be unreliable due to the shifting weight during standing. Moreover, force-based systems fail to provide detailed information about the sub-phases within the swing phase, limiting their utility in comprehensive gait analysis. To overcome these drawbacks, advancements in wearable sensor technology have paved the way for ambulatory gait monitoring and gait detection systems. These systems typically employ accelerometers and gyroscopes to analyze and detect gait events. By utilizing these wearable sensors, researchers and clinicians can obtain more accurate and reliable data for gait analysis. The integration of accelerometers and gyroscopes in gait analysis systems brings forth numerous advantages over traditional force-based sensors. Wearable sensors, such as accelerometers and gyroscopes, exhibit enhanced reliability due to their reduced susceptibility to mechanical failures. Kotiadis et.al [12] This characteristic ensures long-term functionality and accuracy in gait analysis.

One key advantage of wearable sensors is their ability to accommodate the shifting weight of patients with drop foot during standing. This adaptability allows for precise gait event detection in challenging scenarios, improving the accuracy of diagnoses and treatment plans for individuals with gait abnormalities. Moreover, wearable sensors provide valuable insights into the sub-phases of the swing phase, offering a more comprehensive understanding of the intricate mechanics involved in the gait cycle. By capturing detailed data on the different stages within the swing phase, clinicians and researchers gain a deeper understanding of gait patterns and can make more informed decisions regarding treatment and rehabilitation strategies. Recent studies have explored the potential of inertial sensors, including isolated accelerometers, gyroscopes, and inertial measurement units (IMUs), for real-time gait event detection. In comparison to force-based sensors, inertial sensors offer distinct advantages. However, the use of accelerometers may be limited by vibrations during heel strike, potentially affecting their accuracy in certain situations. On the other hand, gyroscopes provide a significant advantage as they are not influenced by gravity and are tolerant towards imprecise positioning. This flexibility allows them to be placed anywhere along the measurement plane, providing consistent and reliable signals. Nevertheless, the current state of development presents some challenges. Gyroscopes typically require significant power, making them less suitable for mobile ambulant usage where power consumption is a concern. Figueiredo et.al[82] Accelerometers, while offering benefits, are influenced by the continuously changing orientation of the measured body segment relative to the gravity field. Accurate sensor positioning becomes crucial to mitigate this bias and ensure accurate gait event detection.

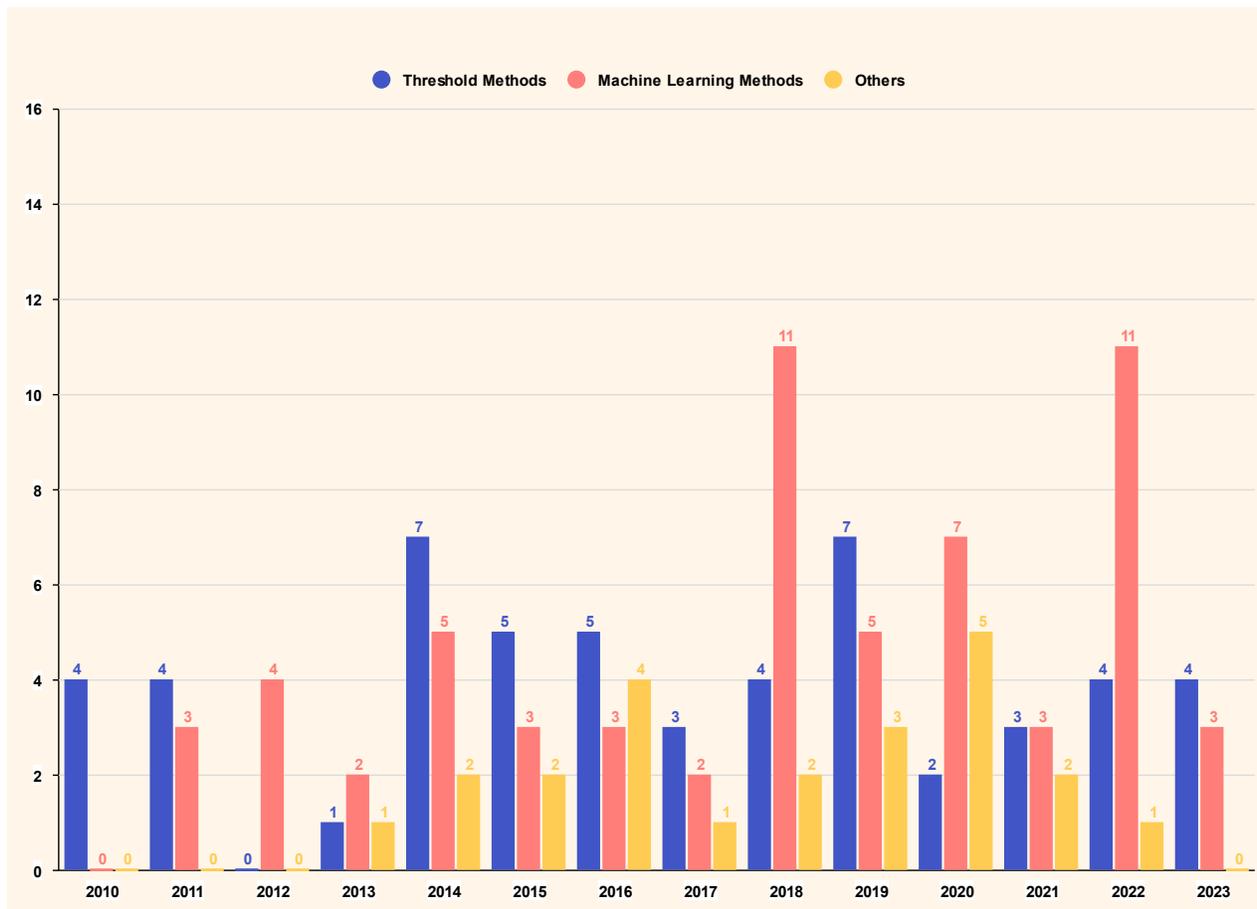

Figure 1. Number of method-based publications from 2010 to 2023

The current state-of-the-art in gait segmentation often relies on heuristic thresholds and rule-based finite state machines (FSMs) when utilizing gyroscope measurements. However, this approach has limitations in terms of time efficiency, reliability, and versatility across different individuals and gait patterns. Therefore, there is a pressing need to develop computational solutions that can accurately describe human gait using a minimal number of wearable sensors, while being easily reproducible in various contexts. To address this challenge, this study aims to review and evaluate a reliable gait event detection system capable of operating during diverse daily locomotion activities. The system under investigation focuses on real-time detection of multiple gait events, including Heel Strike (HS), Foot Flat (FF), Middle Mid-Stance (MMST), Heel-Off (HO), Toe-Off (TO), and Middle Mid-Swing (MMSW).

The evaluation of this system's detection accuracy will be performed using a novel dataset introduced by Camargo et al. in the Journal of Biomechanics. This dataset includes comprehensive information on kinematics, kinetics, and power across various conditions, including walking on a treadmill. By leveraging this dataset and evaluating the performance of the gait event detection system, researchers aim to assess its reliability and accuracy in capturing different gait events during real-life locomotion activities. The study seeks to determine whether the proposed system can provide consistent and precise results across different individuals and gait patterns, thereby establishing its inter-subject and inter-step versatility. This research initiative addresses the demand for a time-effective, reliable, and versatile computational solution for gait analysis. The goal is to enable accurate gait event detection using a minimal number of wearable sensors, making the system easily replicable in different contexts. Ultimately, this study contributes to advancing the field of gait analysis by providing a comprehensive evaluation of a gait event detection system and its potential applicability in both controlled laboratory settings and real-life situations. activities. The goal is to enable accurate gait event detection using a minimal number of wearable sensors, making the system easily replicable in different contexts. Ultimately, this study contributes to advancing the field of gait analysis by providing a comprehensive evaluation of a gait event detection system and its potential applicability in both controlled laboratory settings and real-life situations.

## 2. RELATED WORK

A recent study [12] introduced an efficient algorithm that identifies four key gait events (IC, TO, mid-swing (MSw), and mid-stance (MSt)) by using a single gyroscope attached to the shank of individuals engaged in daily activities like normal walking, fast walking, ramp ascending, and ramp descending. However, this algorithm has certain limitations as it operates offline and does not detect the push-off event, which is a crucial phase preceding toe-off. Although the researchers claim that their algorithm also functions in real-time, they do not provide supporting evidence for their findings. During our literature review, we came across other gait event detection systems that restricted their experiments to a single IMU and categorized gait phases into two [9, 41, 42] or three phases [3]. Within this constrained framework, these studies demonstrated high accuracy rates. For example, authors in [41, 42] achieved a 100% accuracy in detecting IC and TO events. Zhou et al. [21] reported an accuracy above 98% for IC event detection and 95% for TO event detection across three different terrains. We also observed that when the number of phases increased (e.g., four phases [12, 15, 17]), the performance of gait phase detection decreased. For instance, the experiment conducted in [15] yielded an average detection accuracy below 95% for each phase, while Mannini et al. [17] exhibited a significant delay in detection ranging from 45 ms to 100 ms. Furthermore, the experiment conducted in [12] demonstrated a mean difference error of approximately +4 ms for IC and -6.5 ms for TO compared to the reference system.

In a study by Wang et al. [36], the detection of the four primary sub-phases of gait was accomplished by employing an accelerometer attached to the lower portion of the shank. The researchers initially identified the stance phase by monitoring the acceleration in the vertical axis, specifically when it approached 1g. Subsequently, the heel off, swing, and heel strike phases were determined by employing a dynamic threshold derived from the accelerometer data. Numerous control algorithms have been developed using wearable sensors to accurately detect gait events and phases. These techniques involve threshold values, wavelet transformation, and machine learning approaches [105-112]. Rule-based methods (threshold-based approaches) generally offer faster processing times compared to machine learning techniques. Previous studies commonly divide the gait cycle into two main phases by detecting IC and TO, with limited exploration of events occurring within the inner stance phase. However, Mariani et al. [14] conducted research on detecting gait events during the inner stance phase, specifically heel-strike (HS), toe-strike (TS), heel-off (HO), and toe-off (TO), utilizing a single inertial measurement unit (IMU) positioned on the foot. To the authors' knowledge, no other research in the existing literature has devised a real-time gait segmentation approach using only a single-axis gyroscope capable of detecting multiple gait events (such as HS, FF, MMST, HO, TO, and MMSW) in various real-life scenarios available options.

## 3. DATA COLLECTION

The experiment involved a total of 22 healthy participants who were enlisted to perform locomotion trials in four distinct types of terrains, namely a treadmill, level ground, ramp, and stairs. To gather the necessary data, the subjects were equipped with wearable sensors, motion capture markers, and force plates. The wearable sensors were utilized to capture specific movement and biomechanical information from the participants during the trials. These sensors could include devices such as accelerometers, gyroscopes, and inertial measurement units (IMUs), which are commonly attached to various body segments. In addition to the wearable sensors, motion capture markers were strategically placed on the subjects' bodies to track and record their movements with high precision. The motion capture system uses multiple cameras to track the three-dimensional positions of these markers, allowing for accurate motion analysis. The dataset compiled for this study serves as a comprehensive and detailed resource for establishing a characterization of community ambulation across various contexts, including different walking speeds, stairs heights, and ramp inclinations.

It provides valuable information regarding the biomechanics of human locomotion, encompassing a wide range of data sources, including wearable sensors such as IMU (Inertial Measurement Unit), EMG (Electromyography), and goniometers. In addition to capturing the kinematic and kinetic aspects of the participants' movements through motion capture markers and force plates, the dataset includes rich information from wearable sensors. Specifically, it encompasses signals from 8 EMGs, 6 goniometers, 8 IMUs, and 2 force plates, enabling a comprehensive understanding of the participants' biomechanical responses during locomotion. The inclusion of EMG signals provides insights into muscle activation patterns and contributes to a deeper understanding of the neuromuscular aspects of walking. Goniometers, on the other hand, offer measurements of joint angles, aiding in the analysis of joint kinematics and range of motion. The IMU signals provide valuable information about acceleration, orientation, and angular velocity, offering a detailed perspective on the participants' movement patterns and dynamics. Finally, the data from force plates allows for the assessment of ground reaction forces and moments, enabling the analysis of kinetics and power during walking. The IMUs (Inertial Measurement Units) were employed to collect essential data on the angular velocity and acceleration of both the thigh and shank segments. Specifically, the IMU attached to the shank played a pivotal role in detecting gait phases based on the proposed method. By monitoring the angular velocity and acceleration patterns captured by the shank's IMU, the researchers could accurately identify the different phases of the gait cycle. To facilitate precise motion tracking and analysis, markers were strategically positioned on specific anatomical landmarks. These markers were placed on the lateral side of the knee, the sensor itself, the ankle, the heel, the 1st metatarsal, and the 5th metatarsal. This marker placement allowed for the detailed measurement and recording of joint angles, joint positions, and segment movements during locomotion. The force plates were integrated into the experimental setup to capture ground reaction forces and moments. These force plates were instrumental in calculating key moments such as heel down and toe off. The moments of heel down and toe off provide crucial insights into the timing and coordination of gait events. However, for the moment of heel off, the researchers relied on the data obtained from the markers. By analyzing the position and movement data of the markers, particularly the one placed on the heel, the moment of heel off could be accurately determined. This marker-based approach allowed for a comprehensive assessment of the gait cycle and provided valuable information for the analysis of specific gait events.

To ensure ergonomics, ease of use, and portability, each inertial unit in this study was securely housed within protective cases. These cases were designed to be attached to the feet using adjustable ribbons. This setup offered several advantages for the experimental procedure. Firstly, the use of protective cases provided enhanced safety and durability for the inertial units. It minimized the risk of damage to the sensors and electronics, ensuring the reliability and longevity of the equipment. Moreover, attaching the cases to the feet using adjustable ribbons facilitated convenient donning and doffing of the sensor units. This streamlined the preparation process for participants, enabling efficient and hassle-free setup during data collection sessions. Importantly, this attachment method allowed for secure fastening of the sensors to the body part, minimizing sensor motion relative to the foot. By reducing movement and vibrations of the inertial units, fluctuations in the angular velocity signal of the IMUs were minimized. This ensured a more stable and accurate measurement of foot angular velocity. The alignment of the gyroscopes within the sensor units was specifically designed to enable direct measurement of foot angular velocity along the sagittal plane. This alignment optimized the detection and capture of angular velocity variations during gait. By focusing on the sagittal plane, which corresponds to the forward-backward motion of the foot, the study could obtain precise and relevant measurements related to the dynamics of walking. By emphasizing ergonomics, secure attachment, and alignment optimization, the experimental setup and attachment procedure for the inertial units showcased meticulous attention to detail. This approach ensured the comfort and safety of participants while minimizing any potential interference or artifact in the IMU angular velocity signal. The resulting accurate measurement of foot angular velocity in the sagittal plane provided valuable data for the analysis of gait dynamics, contributing to a comprehensive understanding of human locomotion. By combining data from IMUs, force plates, and marker-based motion capture, this study benefited from a multi-modal approach to capturing and analyzing gait dynamics. The IMUs provided detailed information about the angular velocity and acceleration of the thigh and shank segments, while the force plates allowed for the quantification of ground reaction forces and moments. The markers facilitated precise tracking of joint angles and positions, contributing to a comprehensive understanding of the participants' gait patterns. By leveraging this multi-modal data collection approach, the researchers were able to gain valuable insights into the intricate biomechanics of human locomotion, enabling a more comprehensive analysis of gait phases and events. The integration of IMUs, force plates, and marker data strengthened the study's methodology and laid the foundation for a thorough investigation of gait dynamics and related phenomena.

For the ramp trials, the participants performed walking tasks on inclines of six different angles: 5.2°, 7.8°, 9.2°, 11°, 12.4°, and 18°. The trials encompassed both ascending and descending the ramp, as well as the transitions from and to walking on level ground. Stair trials were conducted at four distinct step heights, adhering to the ADA (Americans with Disabilities Act) accessibility guidelines: 10.16cm (4in), 12.70cm (5in), 15.24cm (6in), and 17.78cm (7in). This ensured a comprehensive exploration of stair ambulation within the specified range of step heights. As seen if Fig.1 this dataset was collected from a healthy person with wearable sensors such as goniometer, electromyography sensors (EMGs), inertial motion unit sensors (IMUs) and force plates.

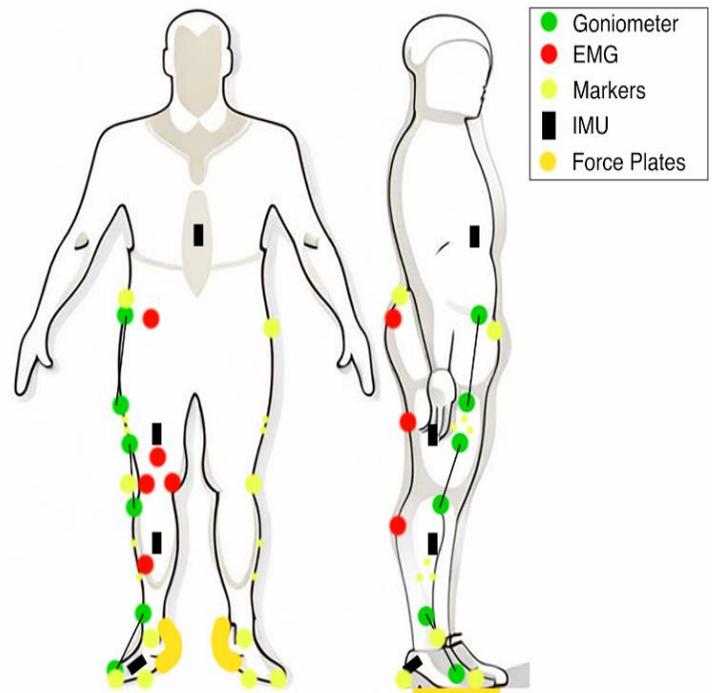

Figure 2. Set-up of wearable sensors in a healthy person

## 4. PERCENT SEGMENTATION AS REFERENCE

The human walking gait is a repetitive sequence involving both legs, starting from the initial contact of one foot with the ground and ending with the subsequent occurrence of the heel of the same foot making contact again. Typically, a gait cycle is divided into two primary phases: the stance phase, which comprises approximately 60% of the gait cycle [43], and the swing phase, which constitutes the remaining 40%. The stance phase begins with the heel-contact event, while the toe-off event marks the start of the swing phase. To provide a finer level of detail, the gait cycle can be further divided into two, three, four, five, six, seven, or eight gait periods, depending on the specific application requirements [35]. These periods define the labels for different stages of the gait cycle [35,43].

Figure 2 illustrates an example of eight periods summarized into one gait cycle, representing 100% of the gait [35,43]. During the initial swing and mid-swing periods, it is crucial to establish the trajectory for the upcoming phase of walking to ensure seamless control of the prosthesis. Prompt detection of gait events is essential to prevent any delays in the device's response. Therefore, an algorithm capable of detecting 100% of the gait cycle is necessary for the advancement of gait phase detection. For applications that do not require a granular breakdown of the gait cycle, the phases can be mapped to the fundamental human gait phases. This mapping facilitates prosthesis control when needed. According to the fundamental human gait phases, the stance phase initiates with the initial contact (IC) occurring from 0% to 10% of the gait cycle. The first 10% of the cycle involves the initial double-limb support phase. The foot flat phase extends from 10% to the point where the heel lifts off the ground at 40% of the gait cycle. Mid-stance occurs at approximately 30% of the gait cycle. Single-limb support extends from foot flat until 50% of the cycle, corresponding to the opposite limb's initial contact.

The second double-limb support phase encompasses the period between the opposite limb's initial contact at 50% and the toe-off event at 60% of the gait cycle. The second single-limb support phase then begins and continues until the completion of the gait cycle. The subsequent periods include early swing (approximately 60–75% of the gait cycle), mid-swing (approximately 75–85% of the gait cycle), and late swing (approximately 85–100% of the gait cycle). These fundamental human gait phases are visually represented in Figure 2. By understanding the fundamental human gait phases and their temporal relationships, researchers and engineers can gain a comprehensive insight into the intricate dynamics of the gait cycle. This knowledge serves as a crucial foundation for the development of advanced algorithms and control strategies, enabling the creation of prosthetic devices that can seamlessly adapt to an individual's gait patterns in real-time. Such advancements have the potential to greatly enhance the functionality and performance of prosthetic robotic systems, ultimately improving the quality of life for individuals relying on assistive wearable technology for gait analysis and state-of-the-art classification methods.

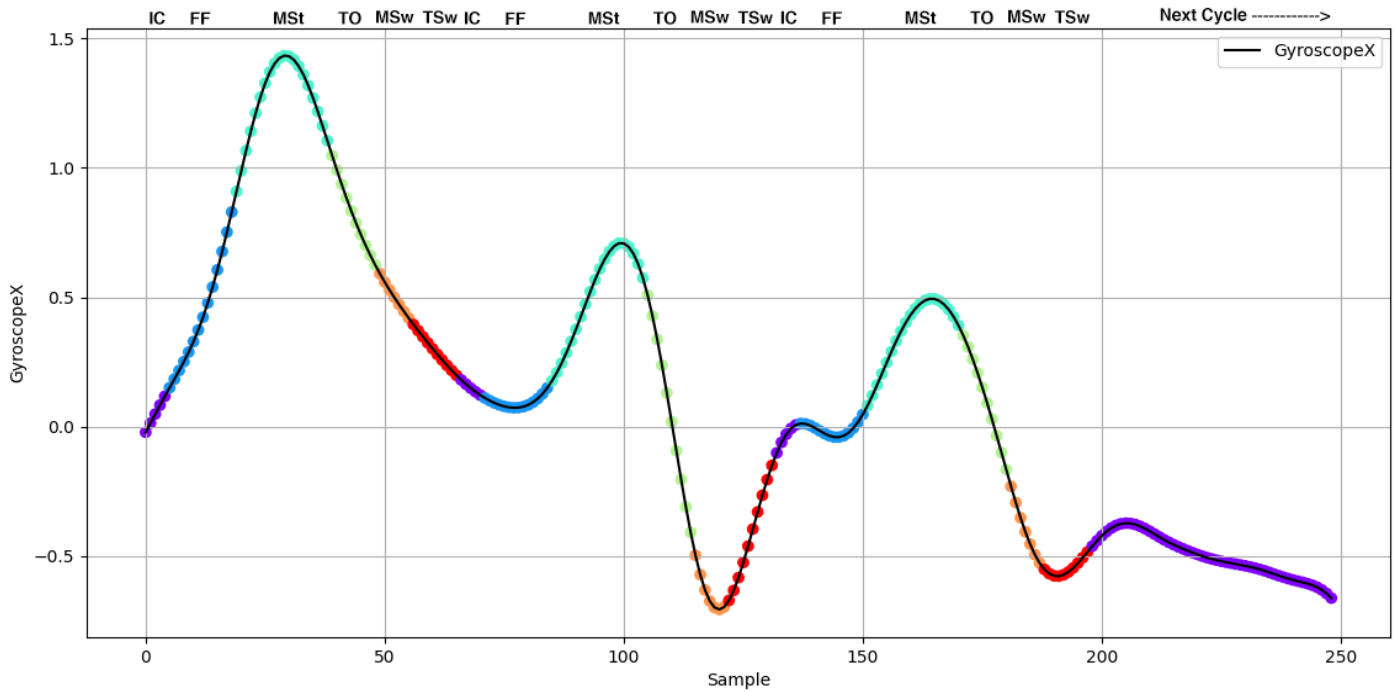

Figure 3. Gait Phases Distribution over a Gait Cycle

## 5. Intelligent Algorithms

The most effective approaches for measuring gait phases and events involve the use of Inertial Measurement Units (IMUs). IMUs encompass gyroscopes, accelerometers, and magnetometers, which capture rotational velocity, linear acceleration, and magnetic field strength along multiple axes. These IMUs can be attached to various body segments such as the foot, shank, or thigh to capture motion signals. Several gait phase classification methods have been developed utilizing acceleration data alone [124,184–188] or gyroscope data [123,135,139,140,142,144,182,189–192]. However, recent research has shown that utilizing both angular velocity and acceleration data from IMUs yields more robust results in detecting gait phases and events [121,133,153,193]. Accelerometers placed on different body segments during gait have demonstrated specific peaks at the beginning and end of the stance phase, enabling the detection of two events using threshold-based or heuristic peak-based methods [124,184]. However, the accuracy of using accelerometers with these methods has shown limitations in specific applications.

On the other hand, gyroscopes have shown promising results, and several studies have proposed their application [123,134,139,140,142,144,183,189–192]. The angular velocity of the joint has been found to be well-suited for Hidden Markov Model (HMM) approaches due to its reliable detection accuracy and ability to detect four phases [27,30,32,71,80]. For instance, Mannini et al. [30] presented an HMM-based system utilizing an IMU placed on the foot to measure angular velocity in the sagittal plane. Their system achieved an average early detection of 45 ms for Foot Flat (FF) phase and a late detection of 35 ms for Heel Off (HO) phase. Another study by Vu et al. [153] proposed a method that utilized all signals from a single IMU placed on the shank. This approach precisely predicted 100% of the gait cycle, with mean square error (MSE) averaging 0.003 in both training and validation datasets.

Simplest computational methods for gait detection involve threshold values. These threshold algorithms consist of a set of rules that determine specific characteristics of gait phases or events [121–125,127,202]. Additionally, time-frequency analysis methods based on thresholding values [130–133,204] and peak heuristic algorithms, which identify points where the derivative passes through zero, are also used [128,131,136,189,205]. The shank's angular velocity signal in the sagittal plane exhibits two distinct negative peaks at the Initial Contact (IC) and Foot Off (FO) events, making it a straightforward rule for accurately detecting these events [123,131,191,192]. Detection success rates have been reported as over 93% [191] and over 98% [123]. In terms of different terrains, the detection reliability of IC and FO events during walking was found to be 95% for Toe Off (TO) and 99% for IC in the upstairs scenario, and 99% for TO and 98% for IC in the downstairs scenario [123,191].

Mariani et al. [14] proposed a method utilizing a single IMU placed on the foot to detect the common stance phase and its sub-phases, including Loading Response, Foot Flat, and Push-Off. The peak method was employed for phase detection, demonstrating reliable precision that could be implemented in real-time applications. Building upon this work, Maqbool and colleagues developed an algorithm based on heuristic rules, which exhibited improved performance compared to previous studies [121]. The mean difference error between the proposed algorithm and the reference system was approximately +4 ms for Heel Contact (HC) and -6.5 ms for TO [128].

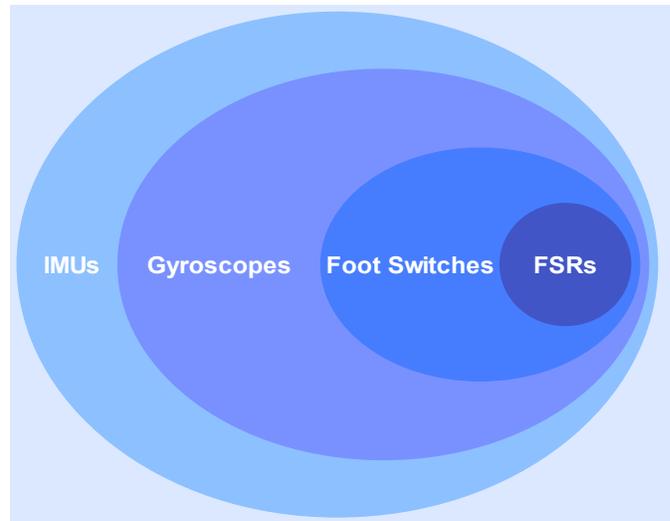

Figure 4. Common Sensors Used in Gait Analysis

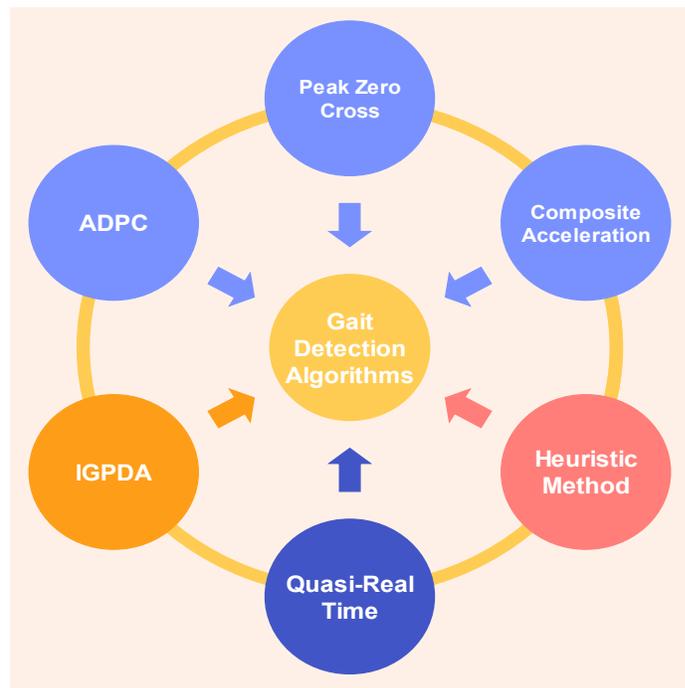

Figure 5: Gait Detection Algorithm discussed in this study

## A. Peak - Zero Cross Method

Yi et al.[86] introduced a novel approach to detect normal and abnormal heel-strike, toe-strike, and toe-off using a single IMU attached to the shank. They successfully identified distinct waveform patterns in the shank's angular velocity and acceleration signals, enabling the differentiation between normal and abnormal gaits. The proposed method employs simple comparators, avoiding complex calculations. Real-time detection of some heel and toe strikes was achieved with a mean absolute error of approximately 2 sample differences, equivalent to a 20ms difference. The Gait-phase classification of this algorithm in our dataset is shown in Figure 6, 7 and 8.

The study revealed unique and shared waveform patterns in the shank's angular velocity and acceleration signals, facilitating the classification of normal and abnormal heel strikes and toe strikes. These findings offer valuable insights into the waveform patterns of the shank's angular velocity, which can be easily comprehended by clinicians to assess patients' gait characteristics. The Implementation of this algorithm is shown in Figure 9, 10 and 11.

In future investigations, the authors plan to extend their method to detect other gait sub-phases, particularly heel-off. Additionally, they intend to evaluate the proposed approach across different walking speeds. Remarkably, the proposed method demonstrates high accuracy in identifying the four types of gait patterns. The detected gait sub-phases, including heel strike, toe strike, and toe-off, exhibit a low mean absolute difference of approximately 2 sample differences.

Although the method may not be suitable for real-time applications, it holds significant promise for gait rehabilitation and other applications that do not necessitate immediate detection. The accuracy of its detection of gait phases is discussed in Figure 28.

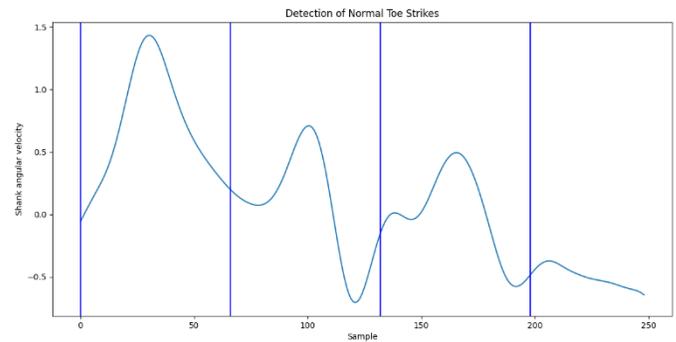

Figure 7. Toe-Strike detections in gait cycles

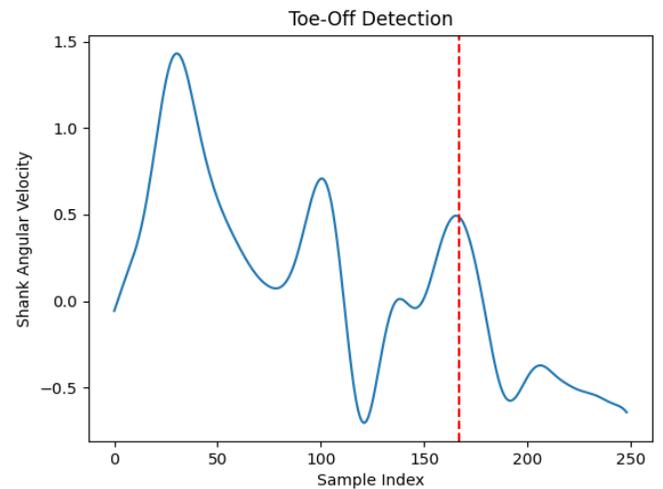

Figure 8. Toe-Off detections in gait cycles

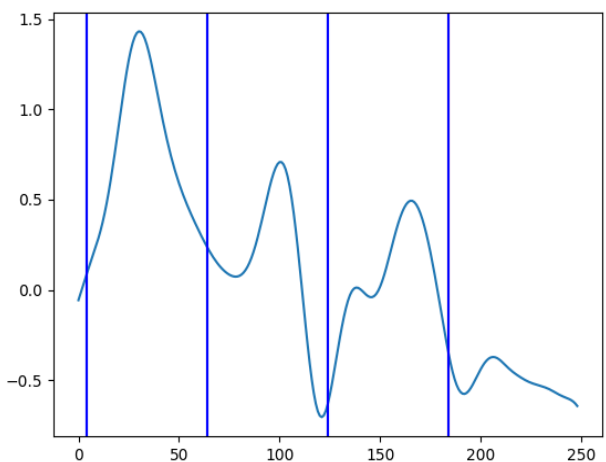

Figure 6. Heel-Strike detections in gait cycles

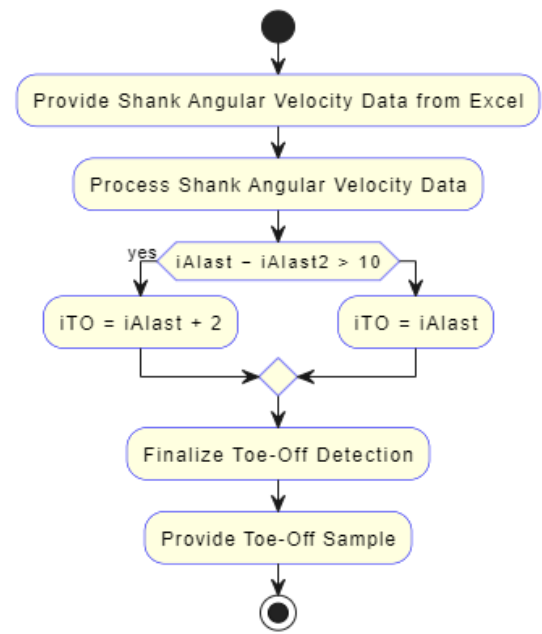

Fig 9. Peak – Zero Cross Toe-off Implementation

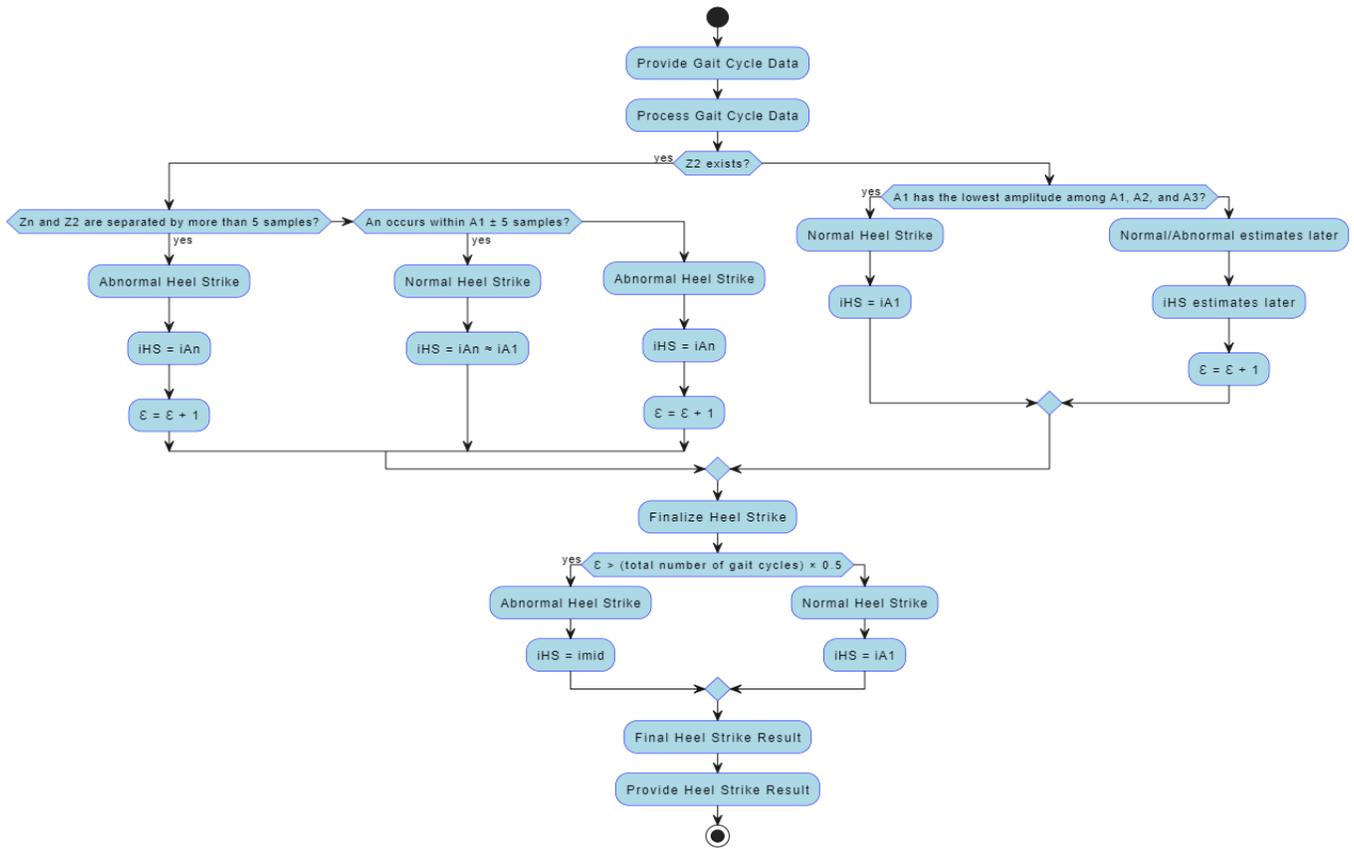

Fig 10. Peak – Zero Cross-Heel-Strike Implementation

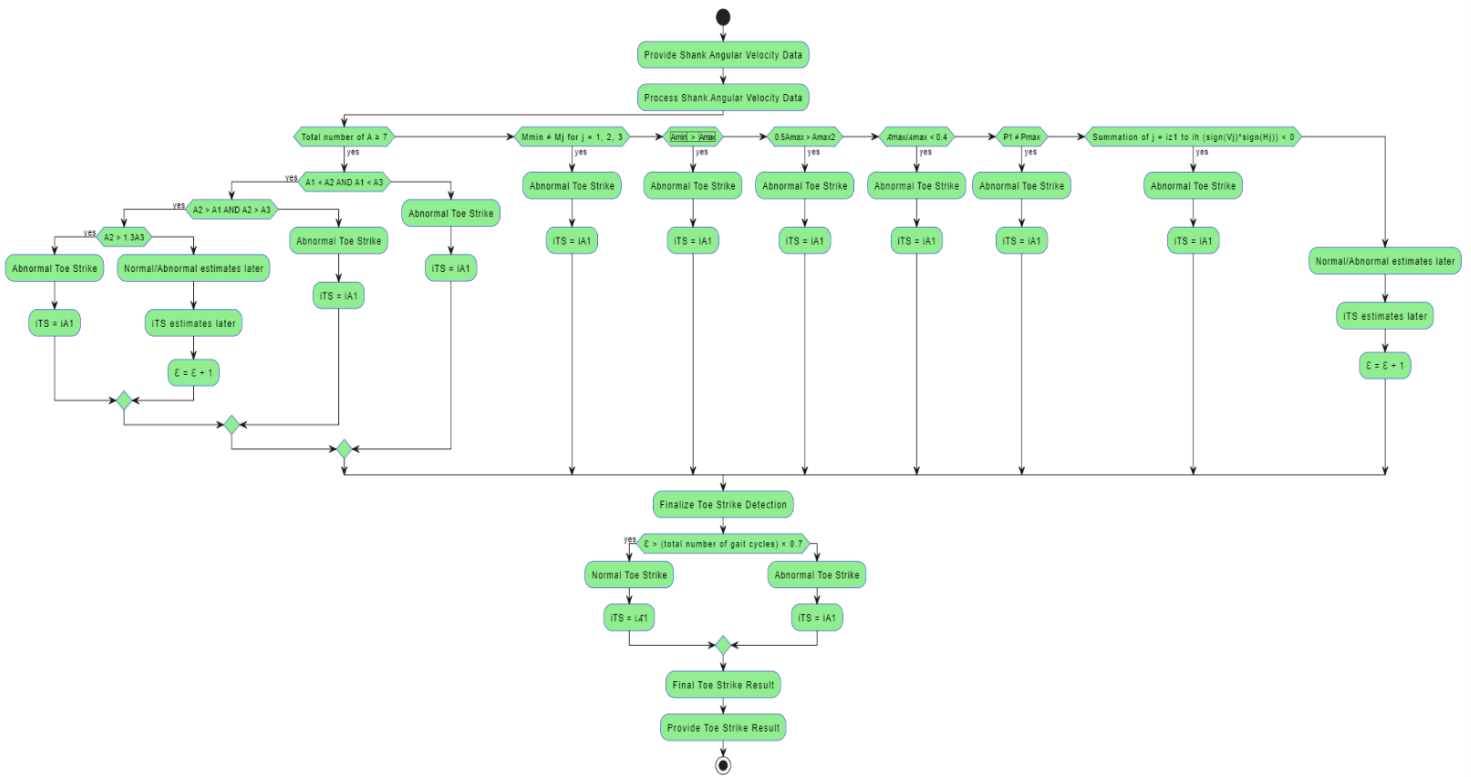

Fig 11. Peak – Zero Cross-Toe-Strike Implementation

## B. Composite Acceleration Method

Rueterbories et al. presented an inertial gait detection algorithm based on a rule-based state machine. This algorithm processes input signals derived from the vectorial sum of radial and tangential acceleration components to generate discrete gait phases: loading response (LR), mid-stance (MS), pre-swing (PS), and swing (SW). The Gait-phase classification of this algorithm in our dataset is shown in Figure 12.

The transitions between gait phases are determined by five reference signals collectively forming the composite acceleration (cA). These reference signals include the low-pass filtered composite acceleration (cA) using a 2nd order Butterworth filter with a cutoff frequency of 6 Hz, as well as its first and second derivatives. Additionally, two low-pass filtered signals of the cA, implemented as moving averages, are used as constraining conditions to detect curve features, particularly during jerky movements. The first filtered signal (cA200) represents the past 1.25 seconds of the gait cycle using a window length of 200 samples, while the second filtered signal (cA50) captures a short history of the previous gait phase with a window length of 50 samples. The use of these filtered signals and derivatives facilitates the detection of inflection points and turning points. The Implementation of this algorithm is shown in Figure 13.

One notable advantage of utilizing angular acceleration is that it is solely generated by rotational motion, eliminating signals resulting from non-rotational motion. Unlike pure accelerometer measures, the angular acceleration signal remains unaffected by gravity. It provides comparable information to gyroscopes but consumes significantly less power. Extracting signal features such as inflection and turning points directly relates to gait phases and requires minimal computational power.

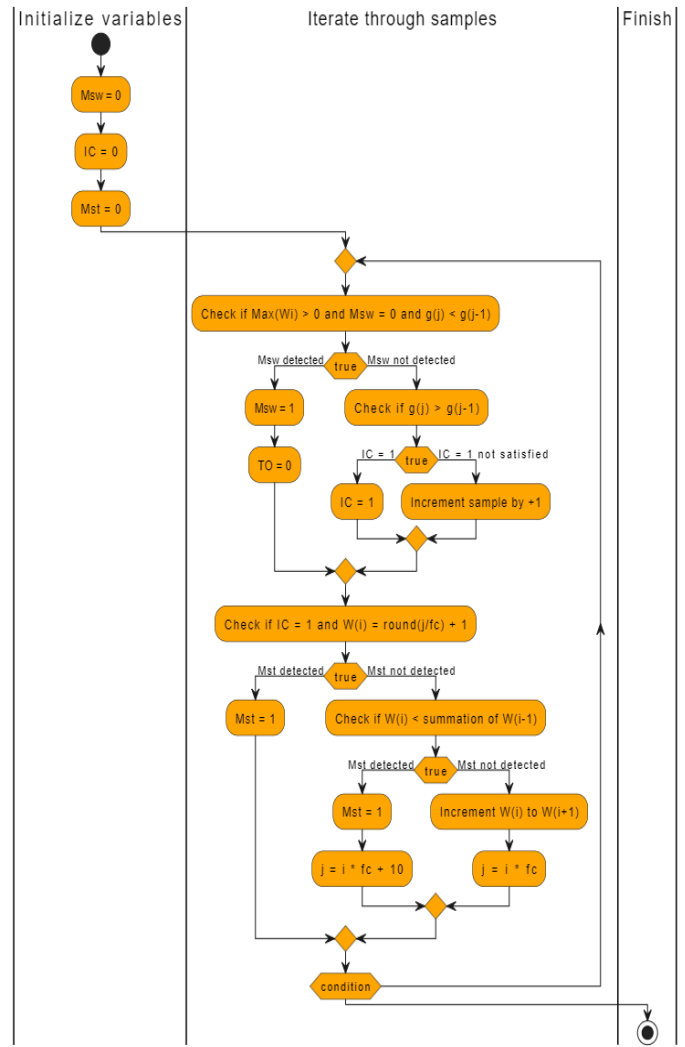

Fig 13. Composite Acceleration Implementation

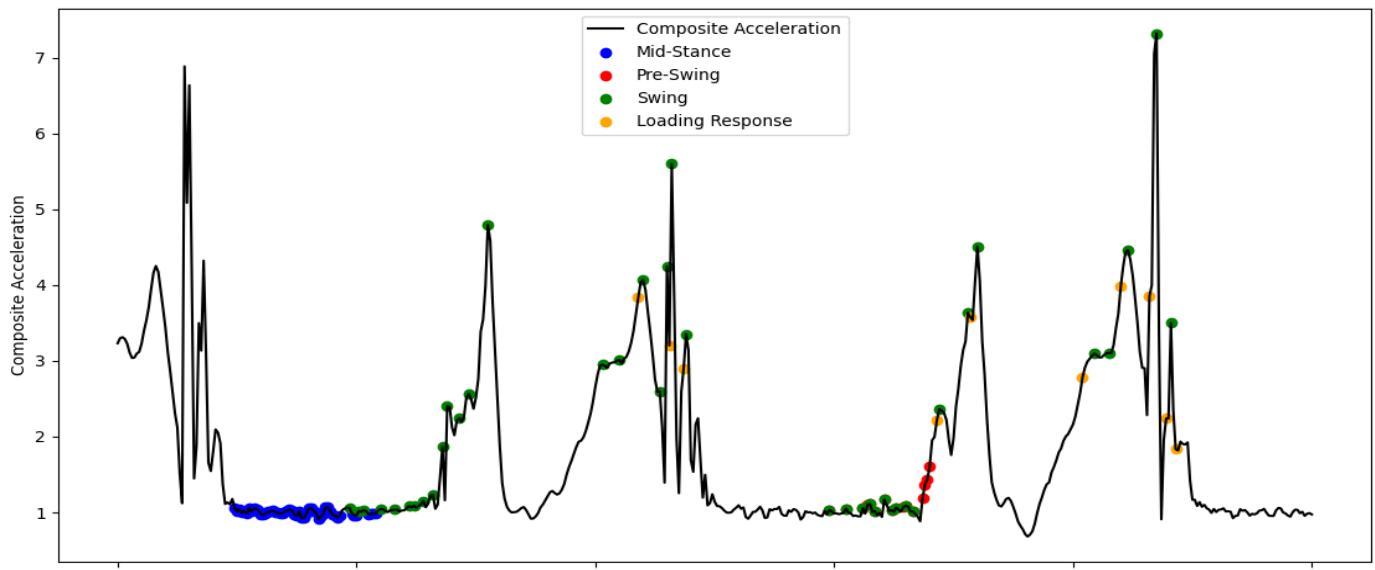

Fig 12. Gait Phase detections using Composite Acceleration Method

## C. Heuristic Method

Zakria et al. introduced a rule-based algorithm for the detection of gait events using a single gyroscope attached to the shank. The algorithm relies on specific criteria to identify the different gait phases: Initial Contact (IC), Toe Off (TO), Mid Stance (MSt), and Mid Swing (MSw). During preprocessing, a 2nd order Butterworth low-pass digital filter with a cutoff frequency of 10 Hz is applied offline to the gyroscope signal. This filtering step reduces oscillations and minimizes the chances of false event detection. The Gait-phase classification of this algorithm in our dataset is shown in Figure 15.

In the algorithm's execution, if the current time (T) is less than a given time (Tg), the algorithm sequentially searches for MSw in the positive slope direction. It detects the occurrence of a change in slope when the current sample (gj) is less than the previous sample (gj-1), indicating a local maximum, which is then labeled as MSw. Once MSw is identified, the algorithm searches for IC in the negative slope direction. It detects a change from negative to positive slope, indicating a local minimum, which is labeled as IC. After marking IC, the algorithm calculates the sum of the samples in the current window (wi) and the previous window (wi-1). If the sum of wi is less than the sum of wi-1, the algorithm marks a peak as MSt. Finally, the algorithm searches for a local minimum peak where the slope changes from negative to positive, marking it as TO. The Implementation of this algorithm is shown in Figure 14.

The proposed system offers a fast and reliable method for gait event detection without the need for threshold values. The mean difference error between the reference system and the gyroscope-based system was found to be approximately +7ms. Future research efforts will focus on evaluating the proposed system with lower limb amputees and in various terrain conditions. The accuracy of its detection of gait phases is discussed in Figure 28.

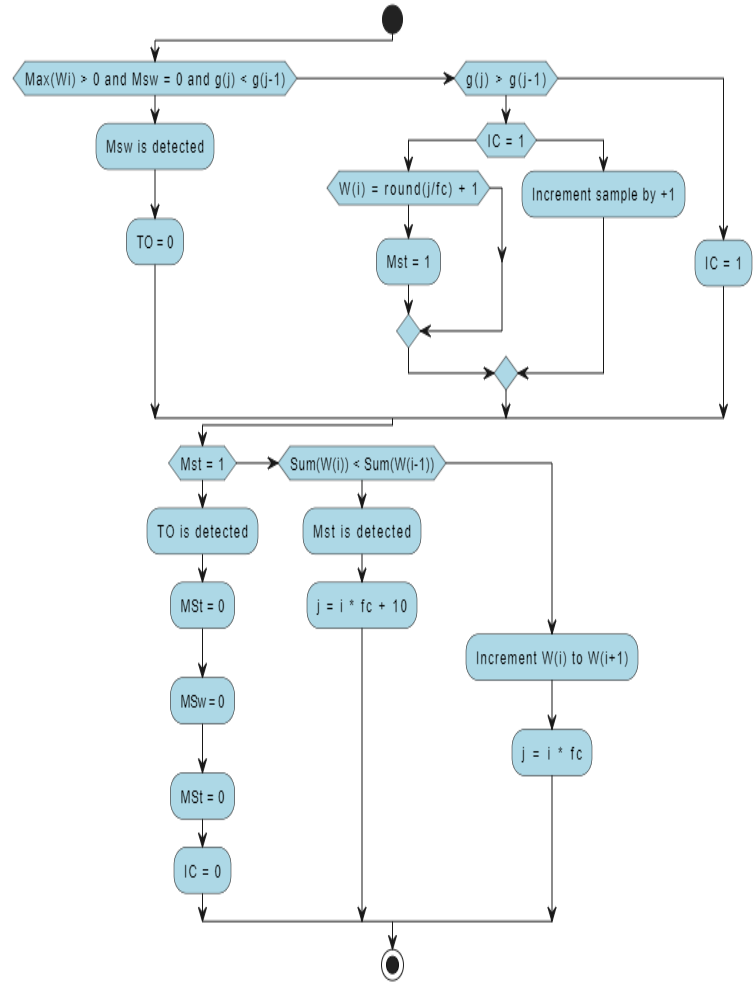

Fig 14. Heuristic Method Implementation

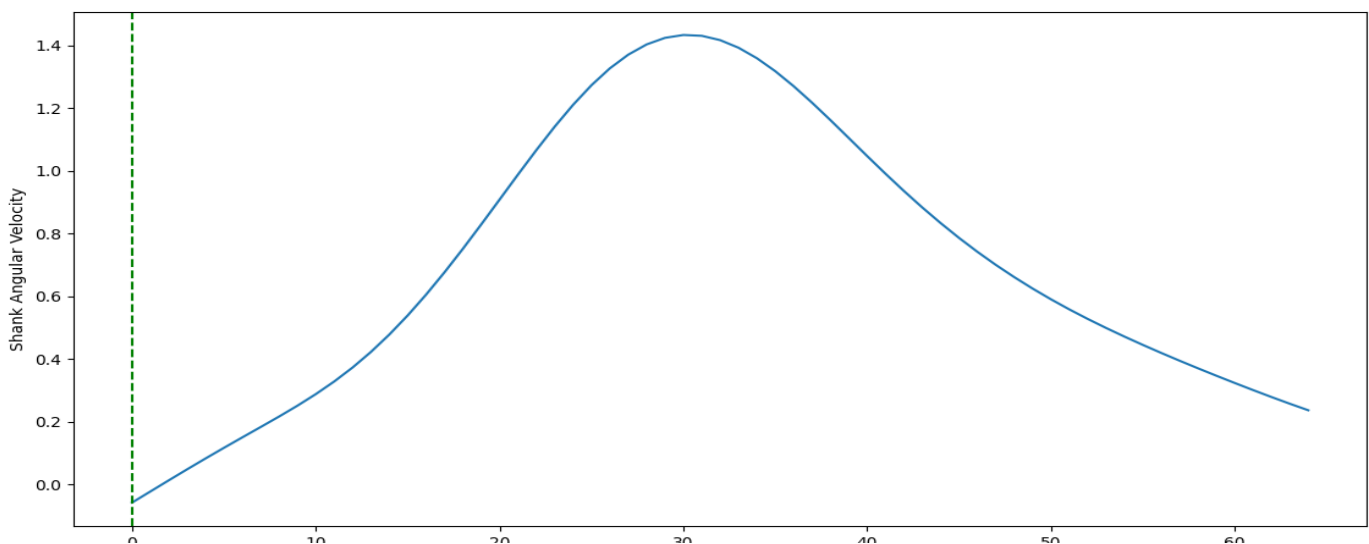

Fig 15. Initial Contact detection using Heuristic Method

## D. Quasi-Real Time Method

Lee et al.[24] proposed an algorithm that focuses on detecting gait events using the mediolateral axis angular velocity from a gyroscope signal attached to the shank. This gyroscope signal exhibits a distinct pattern during the swing phase, characterized by two negative peaks (minima) on either side of a positive peak (maximum). The Gait-phase classification of this algorithm in our dataset is shown in Figure 17.

The algorithm comprises several steps to detect gait events. First, a second-order Butterworth low-pass filter is applied twice to the unfiltered raw signal. This is followed by numerical differentiation to obtain angular acceleration. After these preprocessing steps, the algorithm is divided into three detection parts: mid-swing (MS), event completion (EC), and initial contact (IC). In each part, the algorithm detects events from the 3 Hz low-pass filtered signal, then from the 10 Hz low-pass filtered signal, and finally from the unfiltered raw signal. This approach helps to mitigate erroneous detections caused by signal noise and progressively approach the precise event points. The Implementation of this algorithm is shown in Figure 16.

The proposed algorithm introduces a quasi-real-time approach for automatic gait event detection using uniaxial gyroscopes. It leverages knowledge of the gait event sequence and peak angular acceleration to achieve quasi-real-time detection, which distinguishes it from other algorithms. With a minimal delay (averaged time gap of 0.32 s between EC and MS), the proposed algorithm enables online monitoring of gait events based on gyroscopic measurements. This feature broadens the scope of potential applications beyond post-processing-based algorithms. The accuracy of its detection of gait phases is discussed in Figure 28.

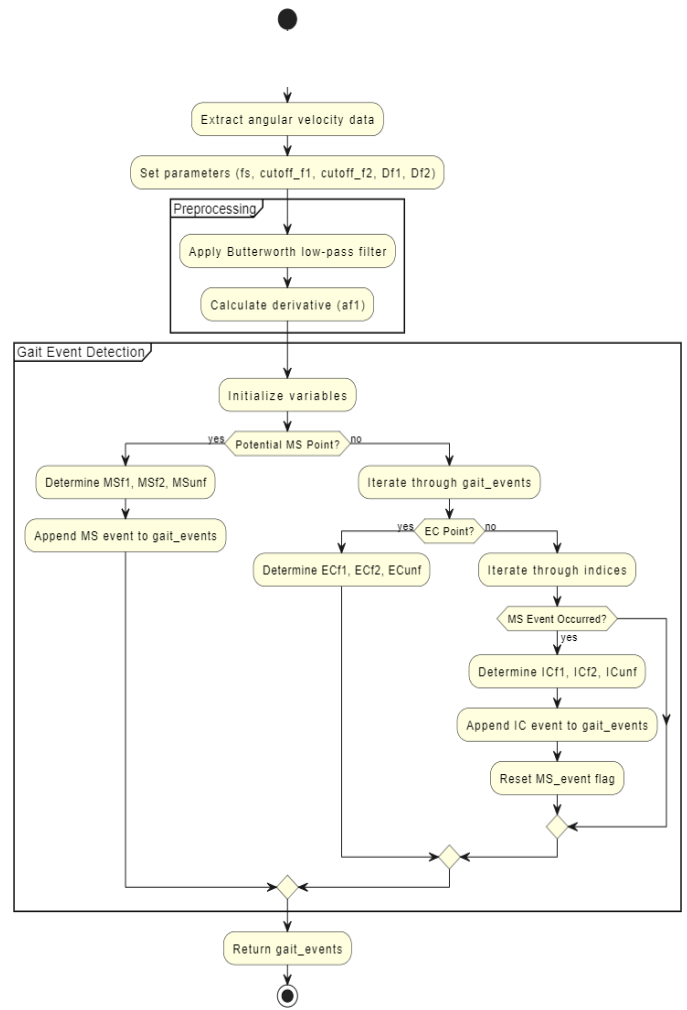

Fig 16. Quasi Real-time Implementation

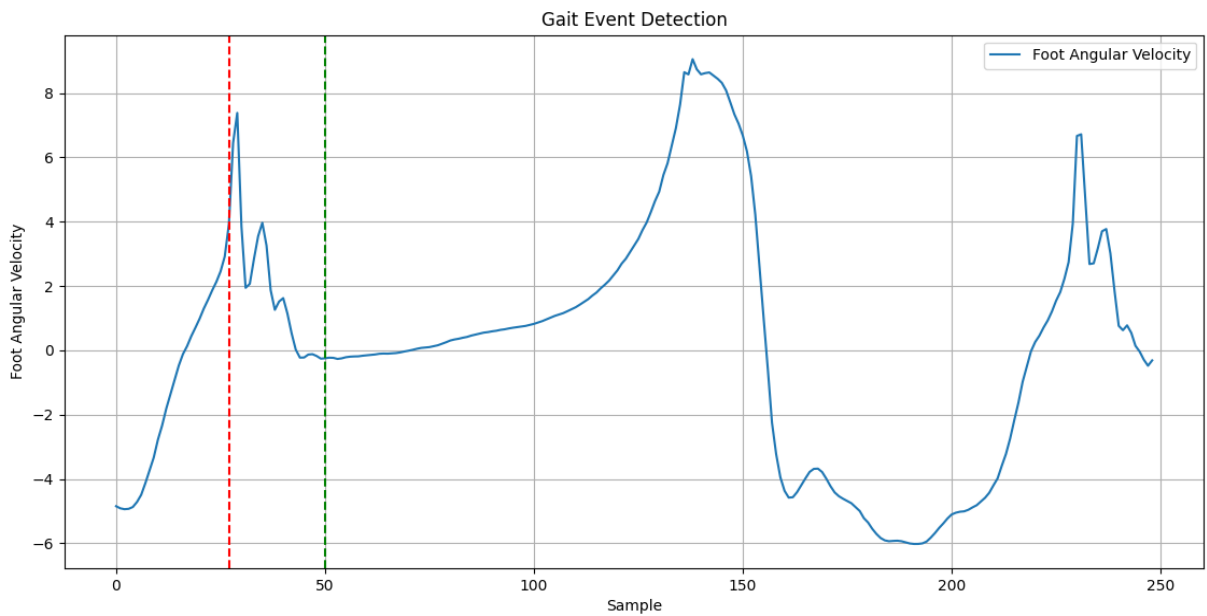

Fig 17. Detection of MSw and IC Phases

## E. Inertial Gait Phase Detection Method

Kotiadis et al. proposed four novel algorithms that utilize data from a 6-DOF inertial sensor to investigate the contribution of different signals to gait phase detection. The first algorithm uses a single accelerometer signal, specifically the radial acceleration (Ax) of the shank. Although the tangential accelerometer (Ay) is more sensitive to shank orientation changes, it also exhibits greater acceleration variations. Therefore, the radial accelerometer (Ax) provides a more reliable representation of shank orientation. The second algorithm incorporates two accelerometers, measuring both radial (Ax) and tangential (Ay) acceleration. The third algorithm utilizes a single gyroscope, measuring angular rotation about the medio-lateral axis (Gz). Finally, the fourth algorithm combines all three components: two accelerometers and one gyroscope. The Gait-phase classification of the IGPDA -1 and IGPDA - 2 respectively in our dataset is shown in Figure 18 and Figure 19 respectively.

To establish a baseline for each accelerometer and gyroscope signal, a first-order recursive low-pass filter was applied, which closely follows the signal average. The filtered signals, denoted as Ax,baseline, Ay,baseline, and Gz,baseline, serve as references for the gait phase detection algorithm. The cut-off frequency for the filter was set to fc = 0.03 Hz. In the case of the heel switch signal, which exhibits large variations, a cut-off frequency of fc = 0.01 Hz was used. The Implementation of the IGPDA -1 and IGPDA - 2 algorithms in our dataset is shown in Figure 20 and Figure 21 respectively.

The individual signals are evaluated against fixed thresholds, which are determined based on the reference baseline of each signal. These thresholds were determined experimentally to optimize performance and are discussed further in the study. By employing thresholds, the system achieves minimum complexity, enabling it to operate with a low computational load. The accelerometer signals undergo low-pass filtering with a cut-off frequency of 10 Hz, utilizing a bi-directional 2nd order (effectively 4th order) Butterworth filter. The accuracy of detection of the IGPDA - 1 and IGPDA - 2 algorithms in our dataset is shown in Figure 28.

All four algorithms were designed in a similar manner, employing the same threshold technique and adhering to identical timing restrictions. These algorithms utilize thresholds derived from the baseline signal to establish conditions for event detection. It is important to note that these thresholds can influence both step detection and timing. The results indicate that the algorithms detected heel off events later than the footswitch. Among the gyroscope algorithms, smaller timing differences with the heel switch were observed compared to the accelerometer algorithms. For heel down events, the algorithms showed earlier detection than the heel switch, except for IGPDA-3 which demonstrated minimal timing differences. The Gait-phase classification of the IGPDA -3 and IGPDA - 4 algorithms in our dataset is shown in Figure 24 and Figure 25 respectively.

The accelerometer algorithms exhibited the detection of weight shifts as steps while failing to detect actual steps. Moreover, a false heel off event was detected during the stance phase. On the other hand, IGPDA-3 and IGPDA-4 demonstrated accurate detection by disregarding weight shifts and successfully identifying steps. The two gyroscope algorithms exhibited effective detection of non-gait events followed by gait events, as demonstrated in the last conjoined step. This highlights the advantages of utilizing shank orientation as a criterion for initiating a step. The Implementation of the IGPDA - 3 and IGPDA - 4 algorithms in our dataset is shown in Figure 22 and Figure 23 respectively.

The research employed a modified MT6-A inertial measurement unit (IMU), provided by Xsens Technologies BV, Enschede, The Netherlands. This IMU consisted of three accelerometers with a 10G accelerometer rating and three gyroscopes with a 900°/s gyroscope rating. This sensor unit facilitated monitoring of all six degrees of freedom, enabling the examination of various sensor combinations at the chosen location. The synchronized data captured by the IMU, along with the inclusion of a heel switch as a reference, ensured comprehensive and accurate measurements. The accuracy of detection of the IGPDA -3 and IGPDA - 4 algorithms in our dataset is shown in Figure 28.

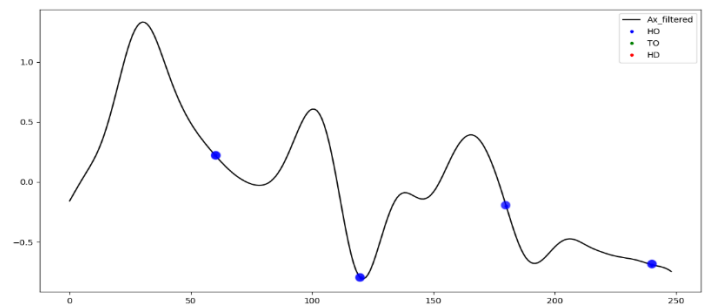

Fig 18. Heel-Off Detection using IGPDA-1

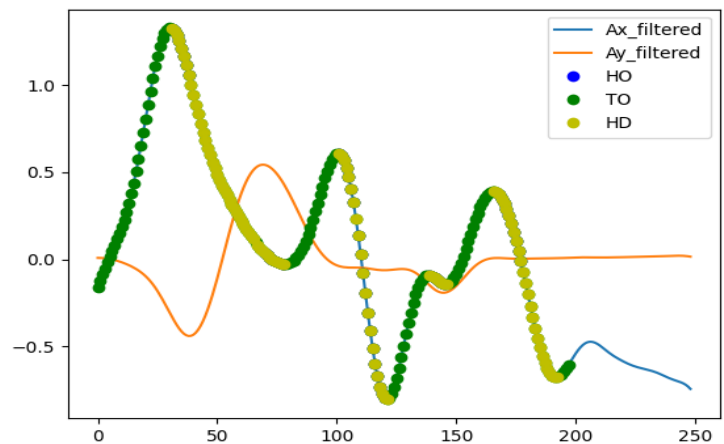

Fig 19. TO and HD Detection using IGPDA-2

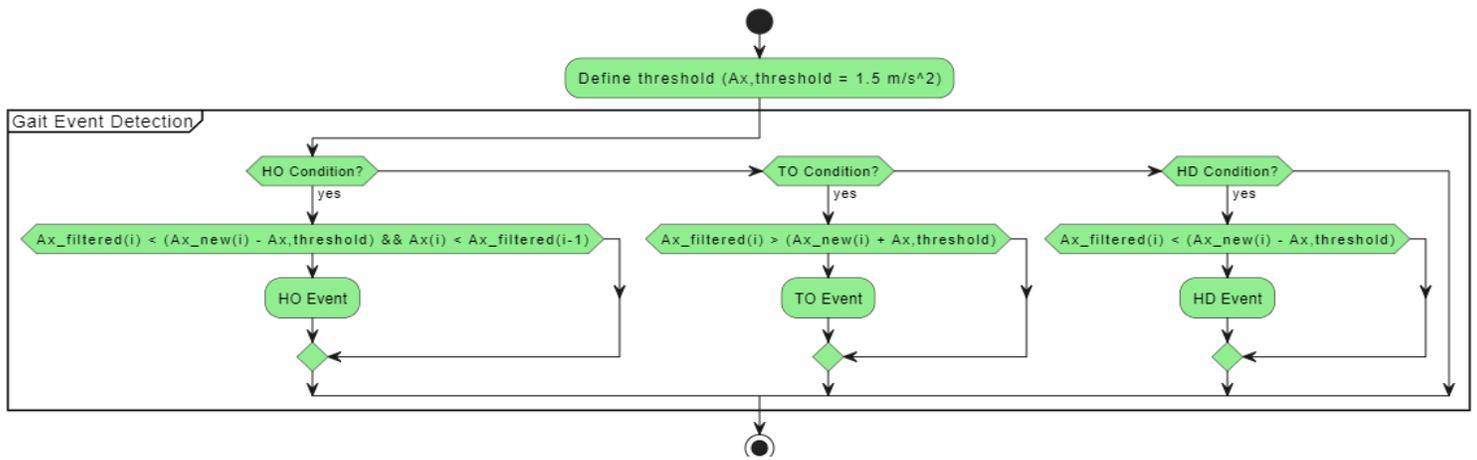

Fig 20. Implementation of IPGDA-1 Algorithm

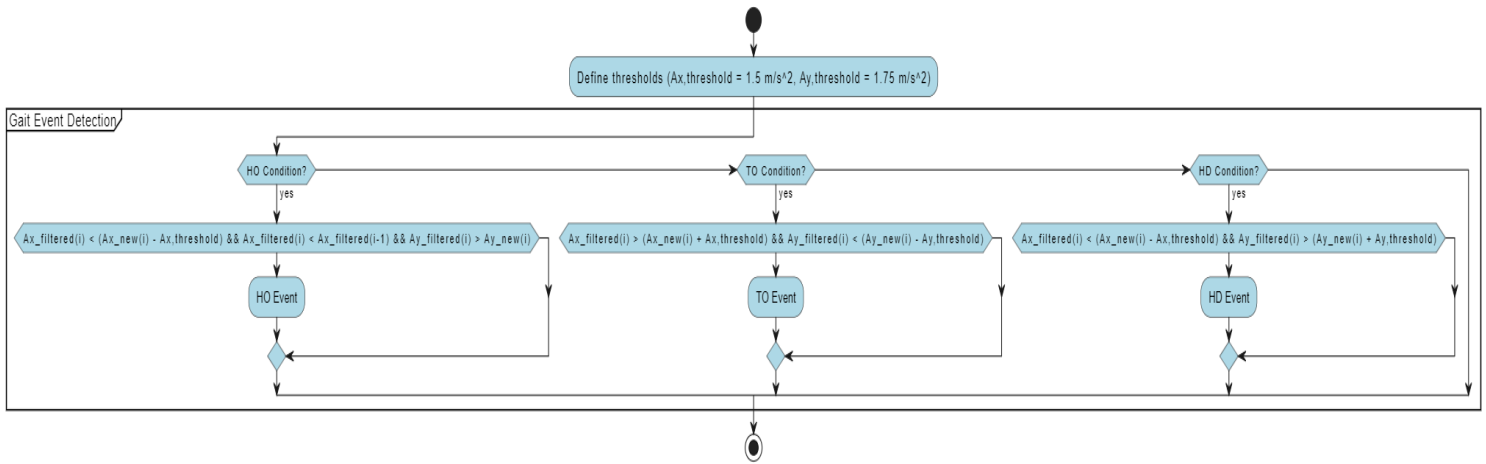

Fig 21. Implementation of IPGDA-2 Algorithm

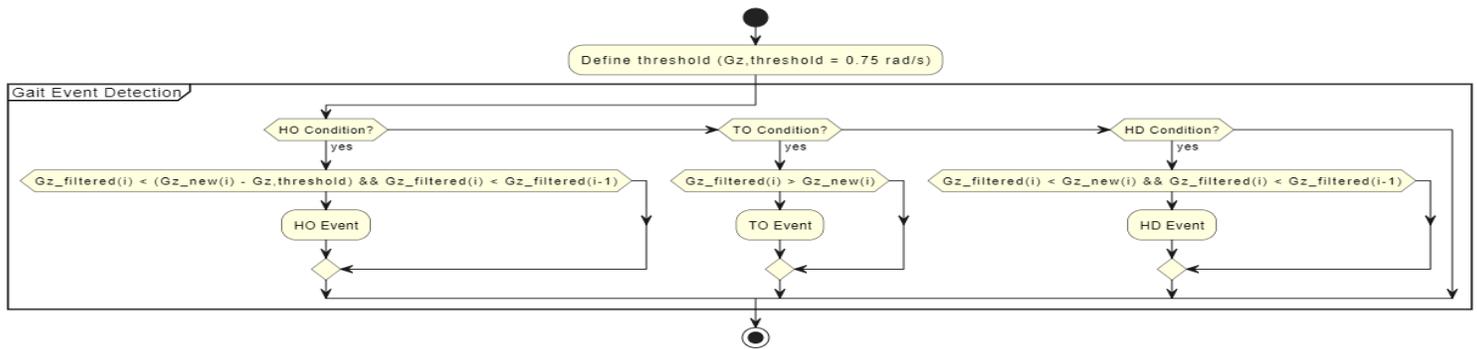

Fig 22. Implementation of IPGDA-3 Algorithm

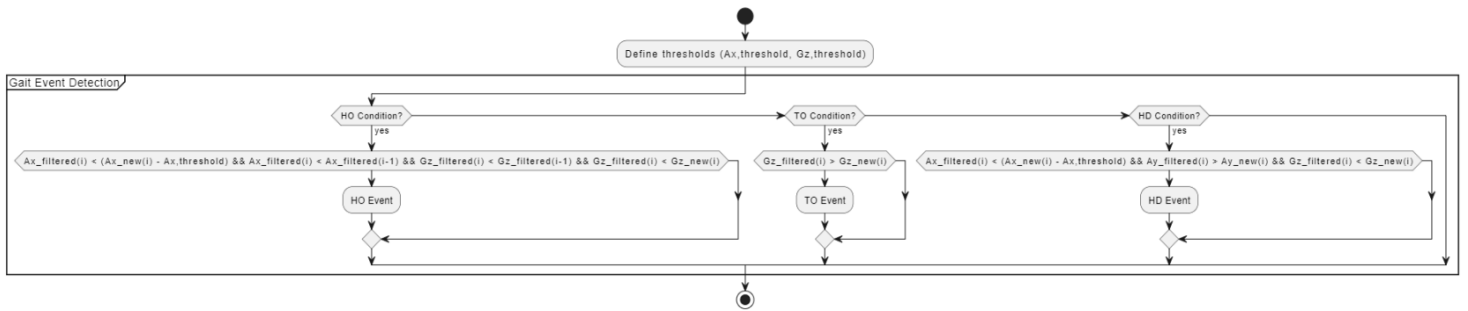

Fig 23. Implementation of IPGDA-4 Algorithm

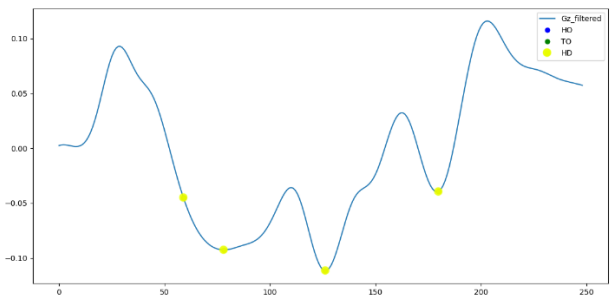

Fig 24. Heel-Down Detection using IGPDA-3

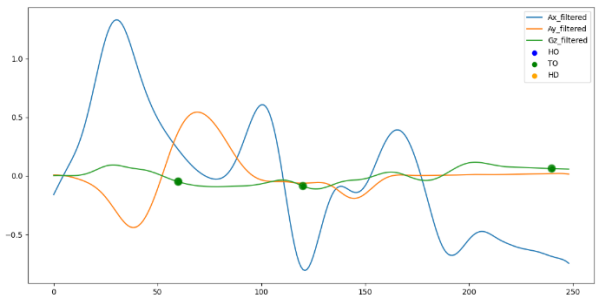

Fig 25. Toe-Off Detection using IGPDA – 4

## F. Adaptive Computational Method

Joana et al.[106] conducted a study verifying that the foot angular velocity exhibited a consistent waveform across various surfaces such as level-ground, inclined surfaces, and staircases. This finding enabled the development of heuristic rules for reliable detection of human gait events. Previous research has demonstrated that this kinematic feature, aligned with the sagittal plane, is versatile for real-time gait detection across different ground surfaces. To define the heuristic decision rules, the angular velocity signal was segmented into six moments corresponding to the six gait events under evaluation: heel strike (HS), foot flat (FF), midstance (MMST), heel off (HO), toe off (TO), and midswing (MMSW). The Gait-phase classification of this algorithm in our dataset is shown in Figure 24.

To establish ground truths for the FF, MMST, and MMSW events, the researchers conducted a visual inspection of video-based angular velocity alongside the IMU angular velocity, with both signals overlapped and synchronized using open-source tracking tools. During the FF and MMST events, when the foot is flat on the ground, the angular velocity remains relatively steady at approximately 0°/s until the HO event occurs, which typically happens after the zero-crossing of the gyroscope signal. It incorporates six different decision rules for the detection of each gait event. These decision rules employ curve tracing techniques, including adaptive threshold crossing, local extrema detection (maximum and minimum angular velocity), and analysis of signal derivatives. In order to account for variations in gait patterns, the proposed method examines changes in the duration and amplitude of angular velocity, as these parameters can vary with different gait speeds. The Implementation of this algorithm is shown in Figure 23.

The first stage involves determining the first derivative by detecting velocity increases (positive signal), decreases (negative signal), or when it becomes approximately zero. The accuracy of its detection of gait phases is discussed in Figure 28.

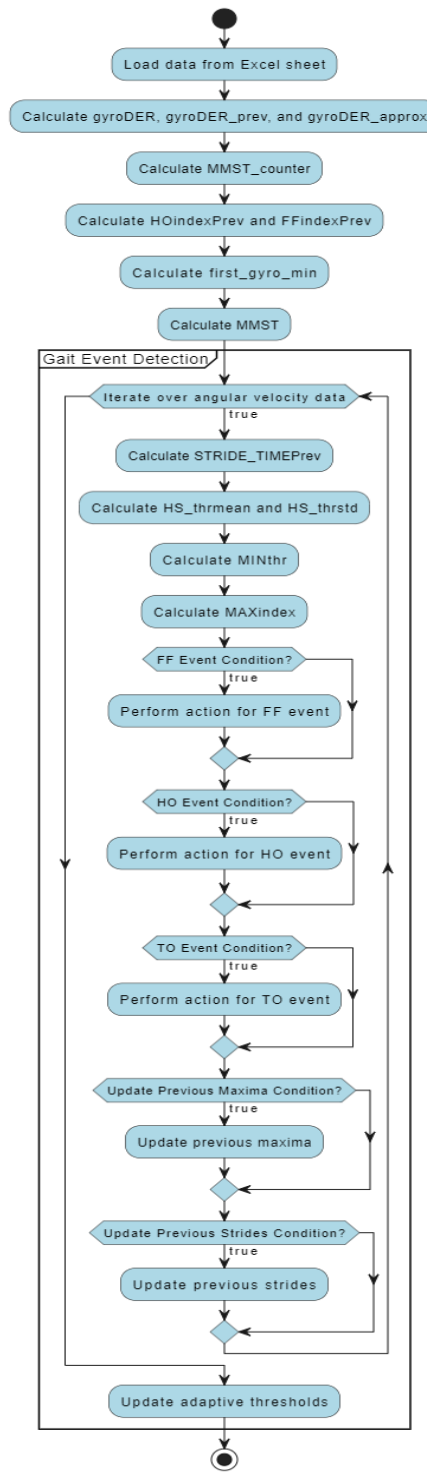

Fig 26. Implementation of Adaptive Computational Algorithm

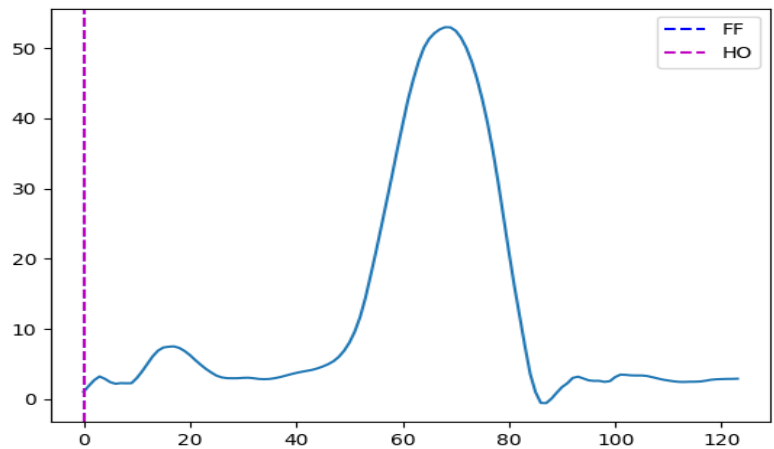

Fig 27. Gait phase detection of Adaptive Computational Algorithm

| Authors | Sensor Types | Placements | Methods | Events Detected | Accuracy |
|---|---|---|---|---|---|
| Yi et. al (2019) | 1 IMU | Shank | Amplitude-Zero cross | HS, TO and TS | ~95% |
| Rueterbories et. al (2014) | Gyroscope | Foot | Composite Acceleration | MSt, PS, Sw, Lr | ~87.5% |
| Zakria et. al (2017) | 1 IMU<br>1 Footswitch | Shank, Foot | Heuristic | IC(unable to detect MSw) | ~50% |
| Lee et. al (2011) | 1 Footswitch<br>1 Gyroscope | Shank and Foot | Quasi Real-Time | MSw and IC<br>(Incorrect detections) | ~50% |
| Kotaidis et. al (2010) | 1 IMUs<br>1 Footswitch | Shank, Foot | IGPDA | HS, HD, HO and TO | ~87.5%<br>(Without footswitch) |
| Joana et. al | 1 Foot Gyroscope | Foot | Adaptive Computational | HO, FF (not detected) | ~50% |

Fig 28. Comparative Analysis of Various Gait Phase Detection Methods

## 8. DISCUSSION

As we go through the different types of gait phase detection algorithms that utilize various versions and implementations of thresholding techniques, we understand that most of the discussed algorithms set parameter thresholds of measures such as foot angular velocity and angular acceleration. This technique, therefore, relies on fine-tuning these parameters and would more accurately work when the subject is subjected to the exact experimental environment and placement of sensors when the algorithm was proposed. Factors such as the anatomical build of the person, activity performed during the data collection, using raw/integrated sensor values, variation in GRF affecting the inverse kinematics parameters, noise, and miscalibration of the sensors are suspected to be the root cause behind the low accuracy of techniques such as [24] and [18].

However, some of these algorithms prove to have much higher accuracies. This study evidently studies the complexity of these algorithms which proves their versatility in various experimental conditions. The composite acceleration method [14] has obtained a nearly 87.5% accuracy as it uses the features of composite accelerations instead of using any one or dual combination of acceleration. This process gives a normalized pattern of the acceleration thus making the probability error due to wrong sensor placement 0 as the signals will consist of all three axes.

[12] also closed an 87.5% accuracy even with the change in the sensor modality this was expected as four separate algorithms were intricately designed using well-defined multiple combinations of various sensor data of each type thus reducing the chances of error. [86] took this to advantage by further creating a pseudo-architecture by using overlapping sliding windows to define a particular combination of Amplitude threshold and a number of zero-cross values for gait phase detection. The adaptive thresholding technique by Joana.et al was also effectively utilized to boost the accuracy of the algorithm close to 95%

## 9. Conclusion

This survey examined numerous studies on state-of-the-art gait phase detection. The review provided a valuable summary of current research and highlighted potential future research directions. It was observed that IMU sensors were commonly used in phase and event detection systems due to their suitability for long-term applications in daily activities. IMUs offered advantages over EMG in terms of energy consumption, durability, cost, weight, portability, and ease of placement. Furthermore, IMU signals could be utilized in all mentioned gait phase detection methods. On the other hand, Gait phase distribution references were useful for experimental reference and system validation. Accuracy rates of 100% were achievable for detecting two phases using any method. However, increasing granularity would lead to a decrease in accuracy within the same gait detection system. This challenge could be addressed through hybrid algorithms, complex algorithms, and incorporating more parameters, which yielded better results. Signal post-processing was crucial to mitigate errors, drift, and signal noise, ensuring high performance. Nonetheless, introducing additional calculation steps could introduce delays, particularly in real-time applications where detection delay is critical. In summary, the choice of detected granularity depends on the requirements of the prosthetic design, considering both detection accuracy and latency.

## References

1. Cherelle, P.; Mathijssen, G.; Wang, Q.; Vanderborght, B.; Lefeber, D. Advances in propulsive bionic feet and their actuation principles. Adv. Mech. Eng. 2014, 6, 984046.

2. Windrich, M.; Grimmer, M.; Christ, O.; Rinderknecht, S.; Beckerle, P. Active lower limb prosthetics: A systematic review of design issues and solutions. Biomed. Eng. Online 2016, 15, 140

3. Dong, D.; Ge, W.; Convens, B.; Sun, Y.; Verstraten, T.; Vanderborght, B. Design, Optimization and Energetic Evaluation of an Efficient Fully Powered Ankle-Foot Prosthesis With a Series Elastic Actuator. IEEE Access

4. 2020, 8, 61491–61503.

5. De Pauw, K.; Cherelle, P.; Roelands, B.; Lefeber, D.; Meeusen, R. The efficacy of the Ankle Mimicking Prosthetic Foot prototype 4.0 during walking: Physiological determinants. Prosthetics Orthot. Int. 2018, 42, 504–510.

6. Flynn, L.; Geeroms, J.; Jimenez-Fabian, R.; Heins, S.; Vanderborght, B.; Munih, M.; Molino Lova, R.; Vitiello, N.; Lefeber, D. The challenges and achievements of experimental implementation of an active transfemoral prosthesis based on biological quasi-stiffness: the CYBERLEGs beta-prosthesis. Front. Neurorobot. 2018, 12, 80.

7. Cherelle, P.; Grosu, V.; Flynn, L.; Junius, K.; Moltedo, M.; Vanderborght, B.; Lefeber, D. The Ankle Mimicking Prosthetic Foot 3—Locking mechanisms, actuator design, control and experiments with an amputee. Robot. Auton. Syst. 2017, 91, 327–336.

8. Taborri, J.; Palermo, E.; Rossi, S.; Cappa, P. Gait partitioning methods: A systematic review. Sensors 2016, 16, 66.

9. Skelly, M.M.; Chizeck, H.J. Real-time gait event detection for paraplegic FES walking. IEEE Trans. Neural Syst. Rehabil. Eng. 2001, 9, 59–68.

10. Maqbool, H.F.; Husman, M.A.B.; Awad, M.I.; Abouhossein, A.; Iqbal, N.; Dehghani-Sanij, A.A. A real-time

11. gait event detection for lower limb prosthesis control and evaluation. IEEE Trans. Neural Syst. Rehabil. Eng. 2017, 25, 1500–1509.

12. Kotiadis, D.; Hermens, H.J.; Veltink, P.H. Inertial gait phase detection for control of a drop foot stimulator. Med. Eng. Phys. 2010, 32, 287–297.

13. Catalfamo, P.; Ghoussayni, S.; Ewins, D. Gait event detection on level ground and incline walking using a rate gyroscope. Sensors 2010, 10, 5683–5702.

14. Rueterbories, J.; Spaich, E.G.; Andersen, O.K. Gait event detection for use in FES rehabilitation by radial and tangential foot accelerations. Med. Eng. Phys. 2014, 36, 502–508.

15. Lau, H.; Tong, K. The reliability of using accelerometer and gyroscope for gait event identification on persons with dropped foot. Gait Posture 2008, 27, 248–257.

16. Behboodi, A.; Wright, H.; Zahradka, N.; Lee, S. Seven phases of gait detected in real-time using shank attached gyroscopes. In Proceeding of the 2015 37th Annual International Conference of the IEEE Engineering in Medicine and Biology Society (EMBC), Milan, Italy, 25–29 August 2015; pp. 5529–5532.

17. Meng, X.; Yu, H.; Tham, M.P. Gait phase detection in able-bodied subjects and dementia patients. In Proceedings of the 2013 35th Annual International Conference of the IEEE Engineering in Medicine and Biology Society (EMBC), Osaka, Japan, 3–7 July 2013; pp. 4907–4910.

18. Zakria, M.; Maqbool, H.F.; Hussain, T.; Awad, M.I.; Mehryar, P.; Iqbal, N.; Dehghani-Sanij, A.A. Heuristic based gait event detection for human lower limb movement. In Proceedings of the 2017 IEEE EMBS International Conference on Biomedical & Health Informatics (BHI), Orlando, FL, USA, 16–19 Febrary 2017; pp. 337–340.

19. Chang, H.C.; Hsu, Y.L.; Yang, S.C.; Lin, J.C.; Wu, Z.H. A wearable inertial measurement system with complementary filter for gait analysis of patients with stroke or Parkinson's disease. IEEE Access 2016, 4, 8442–8453.

20. Ledoux, E. Inertial Sensing for Gait Event Detection and Transfemoral Prosthesis Control Strategy. IEEE Trans. Biomed. Eng. 2018, 65, 2704–2712.

21. Zhou, H.; Ji, N.; Samuel, O.W.; Cao, Y.; Zhao, Z.; Chen, S.; Li, G. Towards Real-Time Detection of Gait Events on Different Terrains Using Time-Frequency Analysis and Peak Heuristics Algorithm. Sensors 2016, 16, 1634.

22. Khandelwal, S.; Wickström, N. Gait event detection in real-world environment for long-term applications: Incorporating domain knowledge into time-frequency analysis. IEEE Trans. Neural Syst. Rehabil. Eng. 2016, 24, 1363–1372.

23. Kim, J.; Bae, M.N.; Lee, K.B.; Hong, S.G. Gait event detection algorithm based on smart insoles. Etri J. 2020, 42, 46–53.

24. Lee, J.K.; Park, E.J. Quasi real-time gait event detection using shank-attached gyroscopes. Med. Biol. Eng. Comput. 2011, 49, 707–712.

25. Goršič, M.; Kamnik, R.; Ambroži c, L.; Vitiello, N.; Lefeber, D.; Pasquini, G.; Munih, M. Online phase detection using wearable sensors for walking with a robotic prosthesis. Sensors 2014, 14, 2776–2794.

26. Grimmer, M.; Schmidt, K.; Duarte, J.E.; Neuner, L.; Koginov, G.; Riener, R. Stance and swing detection based on the angular velocity of lower limb segments during walking. Front. Neurorobotics 2019, 13, 57.

27. Mariani, B.; Rouhani, H.; Crevoisier, X.; Aminian, K. Quantitative estimation of foot-flat and stance phase of gait using foot-worn inertial sensors. Gait Posture 2013, 37, 229–234.

28. Bae, J.; Tomizuka, M. Gait phase analysis based on a Hidden Markov Model. Mechatronics 2011, 21, 961–970.

29. Mannini, A.; Sabatini, A.M. A hidden Markov model-based technique for gait segmentation using a foot-mounted gyroscope. In Proceedings of the 2011 Annual International Conference of the IEEE Engineering in Medicine and Biology Society, Boston, MA, USA, 30 August–3 September 2011; pp. 4369–4373.

30. Evans, R.L.; Arvind, D. Detection of gait phases using orient specks for mobile clinical gait analysis. In Proceedings of the 2014 11th International Conference on Wearable and Implantable Body Sensor Networks, Zurich, Switzerland, 16–19 June 2014; pp. 149–154.

31. Sánchez Manchola, M.D.; Bernal, M.J.P.; Munera, M.; Cifuentes, C.A. Gait phase detection for lower-limb exoskeletons using foot motion data from a single inertial measurement unit in hemiparetic individuals. Sensors 2019, 19, 2988.

32. Mannini, A.; Genovese, V.; Sabatini, A.M. Online decoding of hidden Markov models for gait event detection using foot-mounted gyroscopes. IEEE J. Biomed. Health Inform. 2014, 18, 1122–1130.

34. Sun, W.; Ding, W.; Yan, H.; Duan, S. Zero velocity interval detection based on a continuous hidden Markov model in micro inertial pedestrian navigation. Meas. Sci. Technol. 2018, 29, 065103.

35. Taborri, J.; Scalona, E.; Palermo, E.; Rossi, S.; Cappa, P. Validation of inter-subject training for hidden Markov models applied to gait phase detection in children with cerebral palsy. Sensors 2015, 15, 24514–24529.

36. Zhao, H.; Wang, Z.; Qiu, S.; Wang, J.; Xu, F.; Wang, Z.; Shen, Y. Adaptive gait detection based on foot-mounted inertial sensors and multi-sensor fusion. Inf. Fusion 2019, 52, 157–166.

37. Attal, F.; Amirat, Y.; Chibani, A.; Mohammed, S. Automatic recognition of gait phases using a multiple-regression hidden Markov model. IEEE/ASME Trans. Mechatron. 2018, 23, 1597–1607.

38. Liu, J. Adaptive myoelectric pattern recognition toward improved multifunctional prosthesis control. Med. Eng. Phys. 2015, 37, 424–430.

39. Nazmi, N.; Rahman, M.A.A.; Yamamoto, S.I.; Ahmad, S.A. Walking gait event detection based on electromyography signals using artificial neural network. Biomed. Signal Process. Control. 2019, 47, 334–343.

40. Chen, T.S.; Lin, T.Y.; Hong, Y.W.P. Gait Phase Segmentation Using Weighted Dynamic Time Warping and K-Nearest Neighbors Graph Embedding. In Proceedings of the ICASSP 2020–2020 IEEE International Conference on Acoustics, Speech and Signal Processing (ICASSP), Barcelona, Spain, 4–8 May 2020; pp. 1180–1184.

41. Yang, J.; Huang, T.H.; Yu, S.; Yang, X.; Su, H.; Spungen, A.M.; Tsai, C.Y. Machine Learning Based Adaptive Gait Phase Estimation Using Inertial Measurement Sensors. In Proceedings of the 2019 Design of Medical Devices Conference, Minneapolis, MN, USA, 15–18 April 2019.

42. Chakraborty, S.; Nandy, A. An Unsupervised Approach For Gait Phase Detection. In Proceedings of the 2020 4th International Conference on Computational Intelligence and Networks (CINE), Kolkata, India, 27–29 Febrary 2020; pp. 1–5.

43. Lempereur, M.; Rousseau, F.; Rémy-Néris, O.; Pons, C.; Houx, L.; Quellec, G.; Brochard, S. A new deep learning-based method for the detection of gait events in children with gait disorders: Proof-of-concept and concurrent validity. J. Biomech. 2020, 98, 109490.

44. Vu, H.; Gomez, F.; Cherelle, P.; Lefeber, D.; Nowé, A.; Vanderborght, B. ED-FNN: A new deep learning algorithm to detect percentage of the gait cycle for powered prostheses. Sensors 2018, 18, 2389.

45. Gadaleta, M.; Cisotto, G.; Rossi, M.; Rehman, R.Z.U.; Rochester, L.; Del Din, S. Deep Learning Techniques for Improving Digital Gait Segmentation. In Proceedings of the 2019 41st Annual International Conference of the IEEE Engineering in Medicine and Biology Society (EMBC), Berlin, Germany, 23–27 July 2019; pp. 1834–1837.

46. Di Nardo, F.; Morbidoni, C.; Cucchiarelli, A.; Fioretti, S. Recognition of Gait Phases with a Single Knee Electrogoniometer: A Deep Learning Approach. Electronics 2020, 9, 355.

47. Hannink, J.; Kautz, T.; Pasluosta, C.F.; Gaßmann, K.G.; Klucken, J.; Eskofier, B.M. Sensor-based gait parameter extraction with deep convolutional neural networks. IEEE J. Biomed. Health Inform. 2016, 21, 85–93.

48. Lee, S.S.; Choi, S.T.; Choi, S.I. Classification of gait type based on deep learning using various sensors with smart insole. Sensors 2019, 19, 1757.

49. Agostini, V.; Balestra, G.; Knaflitz, M. Segmentation and classification of gait cycles. IEEE Trans. Neural Syst. Rehabil. Eng. 2014, 22, 946–952.

50. Crea, S.; De Rossi, S.M.; Donati, M.; Reberšek, P.; Novak, D.; Vitiello, N.; Lenzi, T.; Podobnik, J.; Munih, M.; Carrozza, M.C. Development of gait segmentation methods for wearable foot pressure

52. Proceedings of the 2012 4th IEEE RAS & EMBS International Conference on Biomedical Robotics and Biomechanics (BioRob), Rome, Italy, 24–27 June 2012; pp. 361–366

53. Cherelle, P.; Grosu, V.; Matthys, A.; Vanderborght, B.; Lefeber, D. Design and validation of the ankle mimic prosthetic (AMP-) foot 2.0. IEEE Trans. Neural Syst. Rehabil. Eng. 2014, 22, 138–148.

54. Cherelle, P.; Junius, K.; Grosu, V.; Cuypers, H.; Vanderborght, B.; Lefeber, D. The amp-foot 2.1: Actuator design, control and experiments with an amputee. Robotica 2014, 32, 1347–1361.

55. Feng, Y.; Wang, Q. Using One Strain Gauge Bridge to Detect Gait Events for a Robotic Prosthesis. Robotica 2019, 37, 1987–1997.

56. Park, J.S.; Lee, C.M.; Koo, S.M.; Kim, C.H. Gait phase detection using force sensing resistors. IEEE Sens. J. 2020, 20, 6516–6523.

57. Jiang, X.; Chu, K.H.; Khoshnam, M.; Menon, C. A wearable gait phase detection system based on force myography techniques. Sensors 2018, 18, 1279.

58. Moulianitis, V.C.; Syrimpeis, V.N.; Aspragathos, N.A.; Panagiotopoulos, E.C. A closed-loop drop-foot correction system with gait event detection from the contralateral lower limb using fuzzy logic. In Proceedings of the 2011 10th International Workshop on Biomedical Engineering, Kos, Greece, 5–7 October 2011; pp. 1–4.

59. Joshi, C.D.; Lahiri, U.; Thakor, N.V. Classification of gait phases from lower limb EMG: Application to exoskeleton orthosis. In Proceedings of the 2013 IEEE Point-of-Care Healthcare Technologies (PHT), Bangalore, India, 16–18 January 2013; pp. 228–231.

60. Yan, L.; Zhen, T.; Kong, J.L.; Wang, L.M.; Zhou, X.L. Walking Gait Phase Detection Based on Acceleration Signals Using Voting-Weighted Integrated Neural Network. Complexity 2020, 2020.

61. Zhao, Y.; Zhou, S.Wearable device-based gait recognition using angle embedded gait dynamic images and a convolutional neural network. Sensors 2017, 17, 478.

62. Tanghe, K.; Harutyunyan, A.; Aertbeliën, E.; De Groote, F.; De Schutter, J.; Vrancx, P.; Nowé, A. Predicting seat-off and detecting start-of-assistance events for assisting sit-to-stand with an exoskeleton. IEEE Robot. Autom. Lett. 2016, 1, 792–799.

63. Liu, D.X.; Wu, X.; Du, W.; Wang, C.; Xu, T. Gait phase recognition for lower-limb exoskeleton with only joint angular sensors. Sensors 2016, 16, 1579.

sensors. In Proceedings of the 2012 Annual International Conference of the IEEE Engineering in Medicine and Biology Society, San Diego, CA, USA, 28 August–1 September 2012; pp. 5018–5021.

51. 49. De Rossi, S.M.; Crea, S.; Donati, M.; Reberšek, P.; Novak, D.; Vitiello, N.; Lenzi, T.; Podobnik, J.; Munih, M.; Carrozza, M.C. Gait segmentation using bipedal foot pressure patterns. In Proceedings of the 2012 4th IEEE RAS & EMBS International Conference on Biomedical Robotics and Biomechatronics (BioRob), Rome, Italy, 24–27 June 2012; pp. 361–366

64. Sheng, W.; Guo, W.; Zha, F.; Jiang, Z.; Wang, X.; Zhang, H. The Effectiveness of Gait Event Detection Based on Absolute Shank Angular Velocity in Turning. In Proceedings of the 2019 IEEE 4th International Conference on Advanced Robotics and Mechatronics (ICARM), Toyonaka, Japan, 3–5 July 2019; pp. 899–904.

65. Teufl, W.; Lorenz, M.; Miezal, M.; Taetz, B.; Fröhlich, M.; Bleser, G. Towards inertial sensor based mobile gait analysis: Event-detection and spatio-temporal parameters. Sensors 2019, 19, 38.

66. Alonge, F.; Cucco, E.; D'Ippolito, F.; Pulizzotto, A. The use of accelerometers and gyroscopes to estimate hip and knee angles on gait analysis. Sensors 2014, 14, 8430–8446.

67. Quintero, D.; Lambert, D.J.; Villarreal, D.J.; Gregg, R.D. Real-time continuous gait phase and speed estimation from a single sensor. In Proceedings of the 2017 IEEE Conference on Control Technology and Applications (CCTA), Mauna Lani, HI, USA, 27–30 August 2017; pp. 847–852.

68. Shorter, K.A.; Polk, J.D.; Rosengren, K.S.; Hsiao-Wecksler, E.T. A new approach to detecting asymmetries in gait. Clin. Biomech. 2008, 23, 459–467.

69. Neumann, D.A. Neumann, D.A. Kinesiology of the Musculoskeletal System: Foundations for Physical Rehabilitation, 2nd ed.; Elsevier Health Sciences: Amsterdam, The Netherlands, 2010; p. 636.

70. Waters, R.L.; Mulroy, S. The energy expenditure of normal and pathologic gait. Gait Posture 1999, 9, 207–231.

71. Zmitrewicz, R.J.; Neptune, R.R.; Walden, J.G.; Rogers, W.E.; Bosker, G.W. The effect of foot and ankle prosthetic components on braking and propulsive impulses during transtibial amputee gait. Arch. Phys. Med. Rehabil. 2006, 87, 1334–1339.

72. Tiwari, A.; Joshi, D. An Infrared Sensor-Based Instrumented Shoe for Gait Events Detection on Different Terrains and Transitions. IEEE Sens. J. 2020.

73. Martini, E.; Fiumalbi, T.; Dell'Agnello, F.; Ivani´c, Z.; Munih, M.; Vitiello, N.; Crea, S. Pressure-Sensitive Insoles for Real-Time Gait-Related Applications. Sensors 2020, 20, 1448.

74. Taborri, J.; Rossi, S.; Palermo, E.; Patanè, F.; Cappa, P. A novel HMM distributed classifier for the detection of gait phases by means of a wearable

inertial sensor network. Sensors 2014, 14, 16212–16234.

75. Mo, S.; Chow, D.H. Accuracy of three methods in gait event detection during overground running. Gait Posture 2018, 59, 93–98.

76. Zhen, T.; Yan, L.; Yuan, P. Walking Gait Phase Detection Based on Acceleration Signals Using LSTM-DNN Algorithm. Algorithms 2019, 12, 253.

77. Robberechts, P.; Derie, R.; Berghe, P.V.D.; Gerlo, J.; De Clercq, D.; Segers, V.; Davis, J. Gait Event Detection in Tibial Acceleration Profiles: a Structured Learning Approach. Math. Comput. Sci. 2019, 1910, 13372.

78. Flood, M.W.; O'Callaghan, B.; Lowery, M. Gait Event Detection from Accelerometry using the Teager-Kaiser Energy Operator. IEEE Trans. Biomed. Eng. 2019, 67, 658–666.

79. Wu, R.; Wu, J.; Xiao, W. Gait Detection using a Single Accelerometer. In Proceedings of the 2019 IEEE 15th International Conference on Control and Automation (ICCA), Edinburgh, UK, 16–19 July 2019; pp. 178–183.

80. Gouwanda, D.; Gouwanda, A.A. A robust real-time gait event detection using wireless gyroscope and its application on normal and altered gaits. Med. Eng. Phys. 2015, 37, 219–225.

81. Figueiredo, J.; Felix, P.; Costa, L.; Moreno, J.C.; Santos, C.P. Gait event detection in controlled and real-life situations: Repeated measures from healthy subjects. IEEE Trans. Neural Syst. Rehabil. Eng. 2018, 26, 1945–1956.

82. Formento, P.C.; Acevedo, R.; Ghoussayni, S.; Ewins, D. Gait event detection during stair walking using a rate gyroscope. Sensors 2014, 14, 5470–5485.

83. Taborri, J.; Scalona, E.; Rossi, S.; Palermo, E.; Patanè, F.; Cappa, P. Real-time gait detection based on Hidden Markov Model: is it possible to avoid training procedure? In Proceedings of the 2015 IEEE International Symposium on Medical Measurements and Applications (MeMeA), Turin, Italy, 7 May 2015; pp. 141–145.

84. Sahoo, S.; Saboo, M.; Pratihar, D.K.; Mukhopadhyay, S. Real-Time Detection of Actual and Early Gait Events During Level-Ground and Ramp Walking. IEEE Sens. J. 2020, 20, 8128–8136.

85. Han, Y.C.; Wong, K.I.; Murray, I. Gait phase detection for normal and abnormal gaits using IMU. IEEE Sens. J. 2019, 19, 3439–3448.

86. Senanayake, C.M.; Senanayake, S.A. Computational intelligent gait-phase detection system to identify pathological gait. IEEE Trans. Inf. Technol. Biomed. 2010, 14, 1173–1179.

87. Jung, J.Y.; Heo, W.; Yang, H.; Park, H. A neural network-based gait phase classification method using sensors equipped on lower limb exoskeleton robots. Sensors 2015, 15, 27738–27759.

88. Meng, M.; She, Q.; Gao, Y.; Luo, Z. EMG signals based gait phases recognition using hidden Markov models. In Proceedings of the 2010 IEEE International Conference on Information and Automation, Harbin, China, 20–23 June 2010; pp. 852–856.

89. Huang, H.; Kuiken, T.A.; Lipschutz, R.D.; others. A strategy for identifying locomotion modes using surface electromyography. IEEE Trans. Biomed. Eng. 2008, 56, 65–73

90. Zhang, F.; Liu, M.; Huang, H. Effects of locomotion mode recognition errors on volitional control of powered above-knee prostheses. IEEE Trans. Neural Syst. Rehabil. Eng. 2014, 23, 64–72.

91. Huang, H.; Zhang, F.; Hargrove, L.J.; Dou, Z.; Rogers, D.R.; Englehart, K.B. Continuous locomotion-mode identification for prosthetic legs based on neuromuscular–mechanical fusion. IEEE Trans. Biomed. Eng. 2011, 58, 2867–2875.

92. Liu, M.; Zhang, F.; Huang, H.H. An Adaptive Classification Strategy for Reliable Locomotion Mode Recognition. Sensors 2017, 17, 2020.

93. Ding, S.; Ouyang, X.; Liu, T.; Li, Z.; Yang, H. Gait event detection of a lower extremity exoskeleton robot by an intelligent IMU. IEEE Sens. J. 2018, 18, 9728–9735.

94. Bejarano, N.C.; Ambrosini, E.; Pedrocchi, A.; Ferrigno, G.; Monticone, M.; Ferrante, S. A novel adaptive, real-time algorithm to detect gait events from wearable sensors. IEEE Trans. Neural Syst. Rehabil. Eng. 2014, 23, 413–422.

95. Ji, N.; Zhou, H.; Guo, K.; Samuel, O.W.; Huang, Z.; Xu, L.; Li, G. Appropriate mother wavelets for continuous gait event detection based on time-frequency analysis for hemiplegic and healthy individuals. Sensors 2019, 19, 3462.

96. Storm, F.A.; Buckley, C.J.; Mazzà, C. Gait event detection in laboratory and real life settings: Accuracy of ankle and waist sensor based methods. Gait Posture 2016, 50, 42–46.

97. Boutaayamou, M.; Schwartz, C.; Stamatakis, J.; Denoël, V.; Maquet, D.; Forthomme, B.; Croisier, J.L.; Macq, B.; Verly, J.G.; Garraux, G.; others. Development and validation of an accelerometer-based method for quantifying gait events. Med. Eng. Phys. 2015, 37, 226–232.

98. Maqbool, H.F.; Husman, M.A.B.; Awad, M.I.; Abouhossein, A.; Mehryar, P.; Iqbal, N.; Dehghani-Sanij, A.A. Real-time gait event detection for lower limb amputees using a single wearable sensor. In Proceedings of the 2016 IEEE 38th Annual International Conference of the Engineering in Medicine and Biology Society (EMBC), Orlando, FL, USA, 16–20 August 2016; pp. 5067–5070.

99. Mannini, A.; Sabatini, A.M. Gait phase detection and discrimination between walking–jogging activities using hidden Markov models applied to foot motion data from a gyroscope. Gait Posture 2012, 36, 657–661.

100. Kidziński, Ł.; Delp, S.; Schwartz, M. Automatic real-time gait event detection in children using deep neural networks. PLOS ONE 2019, 14, e0211466.

101. Muller, P.; Steel, T.; Schauer, T. Experimental evaluation of a novel inertial sensor based realtime gait phase detection algorithm. In Proceedings of the Technically Assisted Rehabilitation Conference, Berlin, Germany, 1 January 2015

102. X. Wu, D. X. Liu, M. Liu, C. Chen, and H. Gao, ''Individualized gait pattern generation for sharing lower limb exoskeleton robot,'' IEEE Trans. Autom. Sci. Eng., vol. 15, no. 4, pp. 1459–1470, Oct. 2018, doi: 10.1109/TASE.2018.2841358.

103. A. M. Boudali, P. J. Sinclair, and I. R. Manchester, ''Predicting transitioning walking gaits: Hip and knee joint trajectories from the motion of walking canes,'' IEEE Trans. Neural Syst. Rehabil. Eng., vol. 27, no. 9, pp. 1791–1800, Sep. 2019, doi: 10.1109/TNSRE.2019.2933896.

104. Y. C. Han, K. I. Wong and I. Murray, "Gait Phase Detection for Normal and Abnormal Gaits Using IMU," in IEEE Sensors Journal, vol. 19, no. 9, pp. 3439-3448, 1 May1, 2019, doi: 10.1109/JSEN.2019.2894143.

105. Estimation Algorithms Using Gait Phase Information," in IEEE Transactions on Biomedical Engineering, vol. 59, no. 10, pp. 2884-2892, Oct. 2012. doi: 10.1109/TBME.2012.2212245

106. P. Catalfamo, S. Ghoussayni, and D. Ewins, "Gait event detection on level ground and incline walking using a rate gyroscope," Sensors, vol. 10, pp. 5683-5702, 2010.

107. K. Aminian, B. Najafi, C. Büla, P.-F. Leyvraz, and P. Robert, "Spatio-temporal parameters of gait measured by an ambulatory system using miniature gyroscopes," Journal of biomechanics, vol. 35, pp. 689-699, 2002.

108. D. Gouwanda and A. A. Gopalai, "A robust real-time gait event detection using wireless gyroscope and its application on normal and altered gaits," Medical Engineering & Physics, 2015.

109. E. Tileylioglu and A. Yilmaz, "Application of neural based estimation algorithm for gait phases of above knee prosthesis," in Engineering in Medicine and Biology Society

110. (EMBC), 2015 37th Annual International Conference of the IEEE, 2015, pp. 4820-4823.

111. H. Maqbool, M. Husman, M. Awad, A. Abouhossein, and A. Dehghani-Sanij, "Real-time gait event detection for transfemoral amputees during ramp ascending and descending," in Engineering in Medicine and Biology Society (EMBC), 2015 37th Annual International Conference of the IEEE, 2015, pp. 4785-4788.

112. P. Muller, T. Steel, T. Schauer, "Experimental Evaluation of a Novel Inertial Sensor Based Realtime Gait Phase Detection Algorithm," in Proceedings of the Technically Assisted Rehabilitation Conference, 2015.

113. Cherelle, P.; Mathijssen, G.; Wang, Q.; Vanderborght, B.; Lefeber, D. Advances in propulsive bionic feet and their actuation principles. Adv. Mech. Eng. 2014, 6, 984046.

114. Windrich, M.; Grimmer, M.; Christ, O.; Rinderknecht, S.; Beckerle, P. Active lower limb prosthetics: A systematic review of design issues and solutions. Biomed. Eng. Online 2016, 15

115. Dong, D.; Ge, W.; Convens, B.; Sun, Y.; Verstraten, T.; Vanderborght, B. Design, Optimization and Energetic Evaluation of an Efficient Fully Powered Ankle-Foot Prosthesis With a Series Elastic Actuator. IEEE Access 2020, 8, 61491–61503.

116. De Pauw, K.; Cherelle, P.; Roelands, B.; Lefeber, D.; Meeusen, R. The efficacy of the Ankle Mimicking Prosthetic Foot prototype 4.0 during walking: Physiological determinants. Prosthetics Orthot. Int. 2018, 42, 504–510.

117. Flynn, L.; Geeroms, J.; Jimenez-Fabian, R.; Heins, S.; Vanderborght, B.; Munih, M.; Molino Lova, R.; Vitiello, N.; Lefeber, D. The challenges and achievements of experimental implementation of an active transfemoral prosthesis based on biological quasi-stiffness: the CYBERLEGs beta-prosthesis. Front. Neurorobot. 2018, 12, 80.

118. Cherelle, P.; Grosu, V.; Flynn, L.; Junius, K.; Moltedo, M.; Vanderborght, B.; Lefeber, D. The Ankle Mimicking Prosthetic Foot 3—Locking mechanisms, actuator design, control and experiments with an amputee. Robot. Auton. Syst. 2017, 91, 327–336.

119. Taborri, J.; Palermo, E.; Rossi, S.; Cappa, P. Gait partitioning methods: A systematic review. Sensors 2016, 16, 66.

120. Skelly, M.M.; Chizeck, H.J. Real-time gait event detection for paraplegic FES walking. IEEE Trans. Neural Syst. Rehabil. Eng. 2001, 9, Maqbool, H.F.; Husman, M.A.B.; Awad, M.I.; Abouhossein, A.; Iqbal, N.; Dehghani-Sanij, A.A. A real-time gait event detection for lower limb prosthesis control and evaluation. IEEE Trans. Neural Syst. Rehabil. Eng. 2017, 25, 1500–1509.

121. Catalfamo, P.; Ghoussayni, S.; Ewins, D. Gait event detection on level ground and incline walking using a rate gyroscope. Sensors 2010, 10, 5683–5702.

122. Lau, H.; Tong, K. The reliability of using accelerometer and gyroscope for gait event identification on persons with dropped foot. Gait Posture 2008, 27, 248–257.

123. Behboodi, A.; Wright, H.; Zahradka, N.; Lee, S. Seven phases of gait detected in real-time using shank attached gyroscopes. In Proceeding of the 2015 37th Annual International Conference of the IEEE Engineering in Medicine and Biology Society (EMBC), Milan, Italy, 25–29 August 2015; pp. 5529–5532.

124. Meng, X.; Yu, H.; Tham, M.P. Gait phase detection in able-bodied subjects and dementia patients. In Proceedings of the 2013 35th Annual International

Conference of the IEEE Engineering in Medicine and Biology Society (EMBC), Osaka, Japan, 3–7 July 2013; pp. 4907–4910.

125. Chang, H.C.; Hsu, Y.L.; Yang, S.C.; Lin, J.C.; Wu, Z.H. A wearable inertial measurement system with complementary filter for gait analysis of patients with stroke or Parkinson's disease. IEEE Access 2016, 4, 8442–8453.

126. Ledoux, E. Inertial Sensing for Gait Event Detection and Transfemoral Prosthesis Control Strategy. IEEE Trans. Biomed. Eng. 2018, 65, 2704–2712.

127. Khandelwal, S.; Wickström, N. Gait event detection in real-world environment for long-term applications: Incorporating domain knowledge into time-frequency analysis. IEEE Trans. Neural Syst. Rehabil. Eng. 2016, 24, 1363–1372.

128. Lee, J.K.; Park, E.J. Quasi real-time gait event detection using shank-attached gyroscopes. Med. Biol. Eng. Comput. 2011, 49, 707–712.

129. Goršič, M.; Kamnik, R.; Ambrožič, L.; Vitiello, N.; Lefeber, D.; Pasquini, G.; Munih, M. Online phase detection using wearable sensors for walking with a robotic prosthesis. Sensors 2014, 14, 2776–2794.

130. Grimmer, M.; Schmidt, K.; Duarte, J.E.; Neuner, L.; Koginov, G.; Riener, R. Stance and swing detection based on the angular velocity of lower limb segments during walking. Front. Neurorobotics 2019, 13, 57.

131. Bae, J.; Tomizuka, M. Gait phase analysis based on a Hidden Markov Model. Mechatronics 2011, 21, 961–970.

132. Engineering in Medicine and Biology Society, Boston, MA, USA, 30 August–3 September 2011; pp. 4369–4373.

133. Evans, R.L.; Arvind, D. Detection of gait phases using orient specks for mobile clinical gait analysis. In Proceedings of the 2014 11th International Conference on Wearable and Implantable Body Sensor Networks, Zurich, Switzerland, 16–19 June 2014; pp. 149–154.

134. Sánchez Manchola, M.D.; Bernal, M.J.P.; Munera, M.; Cifuentes, C.A. Gait phase detection for lower-limb exoskeletons using foot motion data from a single inertial measurement unit in hemiparetic individuals. Sensors 2019, 19, 2988.

135. Sun, W.; Ding, W.; Yan, H.; Duan, S. Zero velocity interval detection based on a continuous hidden Markov model in micro inertial pedestrian navigation. Meas. Sci. Technol. 2018, 29, 065103.

136. Taborri, J.; Scalona, E.; Palermo, E.; Rossi, S.; Cappa, P. Validation of inter-subject training for hidden Markov models applied to gait phase detection in children with cerebral palsy. Sensors 2015, 15, 24514–24529.

137. Attal, F.; Amirat, Y.; Chibani, A.; Mohammed, S. Automatic recognition of gait phases using a multiple-regression hidden Markov model. IEEE/ASME Trans. Mechatron. 2018, 23, 1597–1607.

138. Liu, J. Adaptive myoelectric pattern recognition toward improved multifunctional prosthesis control. Med. Eng. Phys. 2015, 37, 424–430.

139. Nazmi, N.; Rahman, M.A.A.; Yamamoto, S.I.; Ahmad, S.A. Walking gait event detection based on electromyography signals using artificial neural network. Biomed. Signal Process. Control. 2019, 47, 334–343.

140. Chen, T.S.; Lin, T.Y.; Hong, Y.W.P. Gait Phase Segmentation Using Weighted Dynamic Time Warping and K-Nearest Neighbors Graph Embedding. In Proceedings of the ICASSP 2020–2020 IEEE International Conference on Acoustics, Speech and Signal Processing (ICASSP), Barcelona, Spain, 4–8 May 2020; pp. 1180–1184.

141. Yang, J.; Huang, T.H.; Yu, S.; Yang, X.; Su, H.; Spungen, A.M.; Tsai, C.Y. Machine Learning Based Adaptive Gait Phase Estimation Using Inertial Measurement Sensors. In Proceedings of the 2019 Design of Medical Devices Conference, Minneapolis, MN, USA, 15–18 April 2019.

142. Chakraborty, S.; Nandy, A. An Unsupervised Approach For Gait Phase Detection. In Proceedings of the 2020 4th International Conference on Computational Intelligence and Networks (CINE), Kolkata, India, 27–29 Febrary 2020; pp. 1–5.

143. Lempereur, M.; Rousseau, F.; Rémy-Néris, O.; Pons, C.; Houx, L.; Quellec, G.; Brochard, S. A new deep learning-based method for the detection of gait events in children with gait disorders: Proof-of-concept and concurrent validity. J. Biomech. 2020, 98, 109490.

144. Vu, H.; Gomez, F.; Cherelle, P.; Lefeber, D.; Nowé, A.; Vanderborght, B. ED-FNN: A new deep learning algorithm to detect percentage of the gait cycle for powered prostheses. Sensors 2018, 18, 2389.

145. Gadaleta, M.; Cisotto, G.; Rossi, M.; Rehman, R.Z.U.; Rochester, L.; Del Din, S. Deep Learning Techniques for Improving Digital Gait Segmentation. In Proceedings of the 2019 41st Annual International Conference of the IEEE Engineering in Medicine and Biology Society (EMBC), Berlin, Germany, 23–27 July 2019; pp. 1834–1837.

146. Di Nardo, F.; Morbidoni, C.; Cucchiarelli, A.; Fioretti, S. Recognition of Gait Phases with a Single Knee Electrogoniometer: A Deep Learning Approach. Electronics 2020, 9, 355.

147. Hannink, J.; Kautz, T.; Pasluosta, C.F.; Gaßmann, K.G.; Klucken, J.; Eskofier, B.M. Sensor-based gait parameter extraction with deep convolutional neural networks. IEEE J. Biomed. Health Inform. 2016, 21, 85–9

148. Su, B.Y.; Wang, J.; Liu, S.Q.; Sheng, M.; Jiang, J.; Xiang, K. A CNN-Based Method for Intent Recognition Using Inertial Measurement Units and Intelligent Lower Limb Prosthesis. IEEE Trans. Neural Syst. Rehabil. Eng. 2019, 27, 1032–1042.

149. Agostini, V.; Balestra, G.; Knaflitz, M. Segmentation and classification of gait cycles. IEEE Trans. Neural Syst. Rehabil. Eng. 2014, 22, 946–952.

150. Crea, S.; De Rossi, S.M.; Donati, M.; Reberšek, P.; Novak, D.; Vitiello, N.; Lenzi, T.; Podobnik, J.; Munih, M.; Carrozza, M.C. Development of gait segmentation methods for wearable foot pressure sensors. In Proceedings of the 2012 Annual International Conference of the IEEE Engineering in Medicine and Biology Society, San Diego, CA, USA, 28 August–1 September 2012; pp. 5018–5021.

151. De Rossi, S.M.; Crea, S.; Donati, M.; Reberšek, P.; Novak, D.; Vitiello, N.; Lenzi, T.; Podobnik, J.; Munih, M.; Carrozza, M.C. Gait segmentation using bipedal foot pressure patterns. In Proceedings of the 2012 4th IEEE RAS & EMBS International Conference on Biomedical Robotics and Biomechatronics (BioRob), Rome, Italy, 24–27 June 2012; pp. 361–366.

152. Cherelle, P.; Grosu, V.; Matthys, A.; Vanderborght, B.; Lefeber, D. Design and validation of the ankle mimicking prosthetic (AMP-) foot 2.0. IEEE Trans. Neural Syst. Rehabil. Eng. 2014, 22, 138–148.

153. Cherelle, P.; Junius, K.; Grosu, V.; Cuypers, H.; Vanderborght, B.; Lefeber, D. The amp-foot 2.1: Actuator design, control and experiments with an amputee. Robotica 2014, 32, 1347–1361.

154. Feng, Y.; Wang, Q. Using One Strain Gauge Bridge to Detect Gait Events for a Robotic Prosthesis. Robotica 2019, 37, 1987–1997.

155. Park, J.S.; Lee, C.M.; Koo, S.M.; Kim, C.H. Gait phase detection using force sensing resistors. IEEE Sens. J. 2020, 20, 6516–6523.

156. Jiang, X.; Chu, K.H.; Khoshnam, M.; Menon, C. A wearable gait phase detection system based on force myography techniques. Sensors 2018, 18, 1279.

157. Moulianitis, V.C.; Syrimpeis, V.N.; Aspragathos, N.A.; Panagiotopoulos, E.C. A closed-loop drop-foot correction system with gait event detection from the contralateral lower limb using fuzzy logic. In Proceedings of the 2011 10th International Workshop on Biomedical Engineering, Kos, Greece, 5–7 October 2011; pp. 1–4.

158. Joshi, C.D.; Lahiri, U.; Thakor, N.V. Classification of gait phases from lower limb EMG: Application to exoskeleton orthosis. In Proceedings of the 2013 IEEE Point-of-Care Healthcare Technologies (PHT), Bangalore, India, 16–18 January 2013; pp. 228–231.

159. Tanghe, K.; Harutyunyan, A.; Aertbeliën, E.; De Groote, F.; De Schutter, J.; Vrancx, P.; Nowé, A. Predicting seat-off and detecting start-of-assistance events for assisting sit-to-stand with an exoskeleton. IEEE Robot. Autom. Lett. 2016, 1, 792–799.

160. Liu, D.X.; Wu, X.; Du, W.; Wang, C.; Xu, T. Gait phase recognition for lower-limb exoskeleton with only joint angular sensors. Sensors 2016, 16, 1579.

161. Sheng, W.; Guo, W.; Zha, F.; Jiang, Z.; Wang, X.; Zhang, H. The Effectiveness of Gait Event Detection Based on Absolute Shank Angular Velocity in Turning. In Proceedings of the 2019 IEEE 4th International Conference on Advanced Robotics and Mechatronics (ICARM), Toyonaka, Japan, 3–5 July 2019; pp. 899–904.

162. Teufl, W.; Lorenz, M.; Miezal, M.; Taetz, B.; Fröhlich, M.; Bleser, G. Towards inertial sensor based mobile gait analysis: Event-detection and spatio-temporal parameters. Sensors 2019, 19, 38.

163. Alonge, F.; Cucco, E.; D'Ippolito, F.; Pulizzotto, A. The use of accelerometers and gyroscopes to estimate hip and knee angles on gait analysis. Sensors 2014, 14, 8430–8446.

164. Quintero, D.; Lambert, D.J.; Villarreal, D.J.; Gregg, R.D. Real-time continuous gait phase and speed estimation from a single sensor. In Proceedings of the 2017 IEEE Conference on Control Technology and Applications (CCTA), Mauna Lani, HI, USA, 27–30 August 2017; pp. 847–852.

165. Shorter, K.; Polk, J.D.; Rosengren, K.S.; Hsiao-Wecksler, E.T. A new approach to detecting asymmetries in gait. Clin. Biomech. 2008, 23, 459–467.

166. Neumann, D.A. Neumann, D.A. Kinesiology of the Musculoskeletal System: Foundations for Physical Rehabilitation, 2nd ed.; Elsevier Health Sciences: Amsterdam, The Netherlands, 2010; p. 636.

167. Waters, R.L.; Mulroy, S. The energy expenditure of normal and pathologic gait. Gait Posture 1999, 9, 207–231.

168. Zmitrewicz, R.J.; Neptune, R.R.; Walden, J.G.; Rogers, W.E.; Bosker, G.W. The effect of foot and ankle prosthetic components on braking and propulsive impulses during transtibial amputee gait. Arch. Phys. Med. Rehabil. 2006, 87, 1334–1339.

169. Tiwari, A.; Joshi, D. An Infrared Sensor-Based Instrumented Shoe for Gait Events Detection on Different Terrains and Transitions. IEEE Sens. J. 2020.

170. Martini, E.; Fiumalbi, T.; Dell'Agnello, F.; Ivanić, Z.; Munih, M.; Vitiello, N.; Crea, S. Pressure-Sensitive Insoles for Real-Time Gait-Related Applications. Sensors 2020, 20, 1448.

171. Taborri, J.; Rossi, S.; Palermo, E.; Patanè, F.; Cappa, P. A novel HMM distributed classifier for the

detection of gait phases by means of a wearable inertial sensor network. Sensors 2014, 14, 16212-16234.

172. Mo, S.; Chow, D.H. Accuracy of three methods in gait event detection during overground running. Gait Posture 2018, 59, 93–98.

173. Zhen, T.; Yan, L.; Yuan, P. Walking Gait Phase Detection Based on Acceleration Signals Using LSTM-DNN Algorithm. Algorithms 2019, 12, 253.

174. Robberechts, P.; Derie, R.; Berghe, P.V.D.; Gerlo, J.; De Clercq, D.; Segers, V.; Davis, J. Gait Event Detection in Tibial Acceleration Profiles: a Structured Learning Approach. Math. Comput. Sci. 2019, 1910, 13372.

175. Flood, M.W.; O'Callaghan, B.; Lowery, M. Gait Event Detection from Accelerometry using the Teager-Kaiser Energy Operator. IEEE Trans. Biomed. Eng. 2019, 67, 658–666.

176. Wu, R.; Wu, J.; Xiao, W. Gait Detection using a Single Accelerometer. In Proceedings of the 2019 IEEE 15th International Conference on Control and Automation (ICCA), Edinburgh, UK, 16–19 July 2019; pp. 178–183.

177. Gouwanda, D.; Gouwanda, A.A. A robust real-time gait event detection using wireless gyroscope and its application on normal and altered gaits. Med. Eng. Phys. 2015, 37, 219–225.

178. Formento, P.C.; Acevedo, R.; Ghoussayni, S.; Ewins, D. Gait event detection during stair walking using a rate gyroscope. Sensors 2014, 14, 5470–5485.

179. Taborri, J.; Scalona, E.; Rossi, S.; Palermo, E.; Patanè, F.; Cappa, P. Real-time gait detection based on Hidden Markov Model: is it possible to avoid training procedure? In Proceedings of the 2015 IEEE International Symposium on Medical Measurements and Applications (MeMeA), Turin, Italy, 7 May 2015; pp. 141–145.

180. Sahoo, S.; Saboo, M.; Pratihar, D.K.; Mukhopadhyay, S. Real-Time Detection of Actual and Early Gait Events During Level-Ground and Ramp Walking. IEEE Sens. J. 2020, 20, 8128–8136.

181. Senanayake, C.M.; Senanayake, S.A. Computational intelligent gait-phase detection system to identify pathological gait. IEEE Trans. Inf. Technol. Biomed. 2010, 14, 1173–1179.

182. Jung, J.Y.; Heo, W.; Yang, H.; Park, H. A neural network-based gait phase classification method using sensors equipped on lower limb exoskeleton robots. Sensors 2015, 15, 27738–27759.

183. Meng, M.; She, Q.; Gao, Y.; Luo, Z. EMG signals based gait phases recognition using hidden Markov models. In Proceedings of the 2010 IEEE International Conference on Information and Automation, Harbin, China, 20–23 June 2010; pp. 852–856.

184. Huang, H.; Kuiken, T.A.; Lipschutz, R.D.; others. A strategy for identifying locomotion modes using surface electromyography. IEEE Trans. Biomed. Eng. 2008, 56, 65–73.

185. Zhang, F.; Liu, M.; Huang, H. Effects of locomotion mode recognition errors on volitional control of powered above-knee prostheses. IEEE Trans. Neural Syst. Rehabil. Eng. 2014, 23, 64–72.

186. Huang, H.; Zhang, F.; Hargrove, L.J.; Dou, Z.; Rogers, D.R.; Englehart, K.B. Continuous locomotion-mode identification for prosthetic legs based on neuromuscular–mechanical fusion. IEEE Trans. Biomed. Eng. 2011, 58, 2867–2875.

187. Liu, M.; Zhang, F.; Huang, H.H. An Adaptive Classification Strategy for Reliable Locomotion Mode Recognition. Sensors 2017, 17, 2020.

188. Ding, S.; Ouyang, X.; Liu, T.; Li, Z.; Yang, H. Gait event detection of a lower extremity exoskeleton robot by an intelligent IMU. IEEE Sens. J. 2018, 18, 9728–9735.

189. Bejarano, N.C.; Ambrosini, E.; Pedrocchi, A.; Ferrigno, G.; Monticone, M.; Ferrante, S. A novel adaptive, real-time algorithm to detect gait events from wearable sensors. IEEE Trans. Neural Syst. Rehabil. Eng. 2014, 23, 413–422.

190. Storm, F.A.; Buckley, C.J.; Mazzà, C. Gait event detection in laboratory and real life settings: Accuracy of ankle and waist sensor based methods. Gait Posture 2016, 50, 42–46. Boutaayamou, M.; Schwartz, C.; Stamatakis, J.; Denoël, V.; Maquet, D.; Forthomme, B.; Croisier, J.L.; Macq, B.; Verly, J.G.; Garraux, G.; others. Development and validation of an accelerometer-based method for quantifying gait events. Med. Eng. Phys. 2015, 37, 226–232

191. Maqbool, H.F.; Husman, M.A.B.; Awad, M.I.; Abouhossein, A.; Mehryar, P.; Iqbal, N.; Dehghani-Sanij, A.A. Real-time gait event detection for lower limb amputees using a single wearable sensor. In Proceedings of the 2016 IEEE 38th Annual International Conference of the Engineering in Medicine and Biology Society (EMBC), Orlando, FL, USA, 16–20 August 2016; pp. 5067–5070.

192. Kidzin´ski, Ł.; Delp, S.; Schwartz, M. Automatic real-time gait event detection in children using deep neural networks. PLOS ONE 2019, 14, e0211466.

193. Tamon Miyake, Yo Kobayashi, Masakatsu G. Fujie & Shigeki Sugano (2020) Gait event detection based on inter-joint coordination using only angular information, Advanced Robotics, 34:18, 1190-1200, DOI: 10.1080/01691864.2020.1803126

194. S. Chen, S. S. Bangaru, T. Yigit, M. Trkov, C. Wang and J. Yi, "Real-Time Walking Gait Estimation for Construction Workers using a Single Wearable Inertial Measurement Unit (IMU)," 2021 IEEE/ASME

International Conference on Advanced Intelligent Mechatronics (AIM), Delft, Netherlands, 2021, pp. 753-758, doi: 10.1109/AIM46487.2021.9517592.

195. Laidig, D., Jocham, A. J., Guggenberger, B., Adamer, K., Fischer, M., & Seel, T. (2021). Calibration-free gait assessment by foot-worn inertial sensors. Frontiers in Digital Health, 3, 736418.

196. Milad Nazarahari, Aminreza Khandan, Atif Khan, Hossein Rouhani,Foot angular kinematics measured with inertial measurement units: A reliable criterion for real-time gait event detection, Journal of Biomechanics,Volume 130,2022,110880,ISSN 0021-9290

197. Zhang, X., Zhang, H., Hu, J., Zheng, J., Wang, X., Deng, J., ... & Wang, Y. (2022). Gait Pattern Identification and Phase Estimation in Continuous Multilocomotion Mode Based on Inertial Measurement Units. IEEE Sensors Journal, 22(17), 16952-16962.

198. F. A. Garcia, J. C. Pérez-Ibarra, M. H. Terra and A. A. G. Siqueira, "Adaptive Algorithm for Gait Segmentation Using a Single IMU in the Thigh Pocket," in IEEE Sensors Journal, vol. 22, no. 13, pp. 13251-13261, 1 July1, 2022, doi: 10.1109/JSEN.2022.3177951.

199. Shibagaki, K., Arai, S., Milosevic, M., & Nomura, T. (2022). Development of a gait phase detection algorithm for finite state control of functional electrical stimulation during walking. IEICE Technical Report; IEICE Tech. Rep., 121(338), 43-43.

200. Kim, G. T., Lee, M., Kim, Y., & Kong, K. (2023). Robust Gait Event Detection Based on the Kinematic Characteristics of a Single Lower Extremity. International Journal of Precision Engineering and Manufacturing, 1-14.

201. Qin, S., Chen, X., Li, P., & Sun, H. (2023). Estimation of Gait Subphase Time Parameters Based on a Human Electrostatic Field Detection System. IEEE Sensors Journal, 23(9), 9716-9726.

202. Park, K. W., Choi, J., & Kong, K. (2023). Data-Driven Modeling for Gait Phase Recognition in a Wearable Exoskeleton Using Estimated Forces. IEEE Transactions on Robotics.

203. D. Hollinger, M. Schall, H. Chen, S. Bass and M. Zabala, "The Influence of Gait Phase on Predicting Lower-Limb Joint Angles," in IEEE Transactions on Medical Robotics and Bionics, vol. 5, no. 2, pp. 343-352, May 2023, doi: 10.1109/TMRB.2023.3260261.